\documentclass[acmsmall]{acmart}

\AtBeginDocument{%
  }

\setcopyright{acmcopyright}
\copyrightyear{2018}
\acmYear{2018}
\acmDOI{XXXXXXX.XXXXXXX}

\acmJournal{TOMM}
\acmVolume{37}
\acmNumber{4}
\acmArticle{111}
\acmMonth{8}

\usepackage{multirow}
\usepackage{pifont}

\usepackage{booktabs}
\usepackage{float}

\usepackage{colortbl}

\newcommand{\ie}{\textit{i.e.} }
\newcommand{\eg}{\textit{e.g.} }
\newcommand{\etal}{\textit{et al.} }

\begin{document}

\title{Contextual Interaction via Primitive-based Adversarial Training for Compositional Zero-shot Learning}

\author{Suyi Li}
\email{lisuyi@njust.edu.cn}
\orcid{0009-0003-4671-7891}
\author{Chenyi Jiang}
\email{jiangchenyi@njust.edu.cn}
\affiliation{%
  \institution{School of Computer Science and Engineering, Nanjing University of Science and Technology}
  \city{Nanjing}
  \state{Jiangsu}
  \country{China}
}

\author{Shidong Wang}
\email{Shidong.wang@newcastle.ac.uk}
\affiliation{%
  \institution{School of Engineering, Newcastle University}
  \city{Newcastle upon Tyne}
  \country{United Kingdom}
}

\author{Yang Long}
\email{yang.long@ieee.org}
\affiliation{%
  \institution{Department of Computer Science, Durham University}
  \city{Durham}
  \country{United Kingdom}
}

\author{Zheng Zhang}
\email{darrenzz219@gmail.com}
\affiliation{%
  \institution{Shenzhen Key Laboratory of Visual Object Detection and Recognition, Harbin Institute of Technology}
  \city{Shenzhen}
  \country{China}
}

\author{Haofeng Zhang}
\email{zhanghf@njust.edu.cn}
\authornote{Corresponding author.}
\affiliation{%
 \institution{School of Artificial Intelligence, Nanjing University of Science and Technology}
  \city{Nanjing}
  \state{Jiangsu}
  \country{China}
}

%
%
%

\renewcommand{\shortauthors}{Li et al.}

\begin{abstract}
 Compositional Zero-shot Learning (CZSL) aims to identify novel compositions via known attribute-object pairs. The primary challenge in CZSL tasks lies in the significant discrepancies introduced by the complex interaction between the visual primitives of attribute and object, consequently decreasing the classification performance towards novel compositions. Previous remarkable works primarily addressed this issue by focusing on disentangling strategy or utilizing object-based conditional probabilities to constrain the selection space of attributes. Unfortunately, few studies have explored the problem from the perspective of modeling the mechanism of visual primitive interactions. Inspired by the success of vanilla adversarial learning in Cross-Domain Few-Shot Learning, we take a step further and devise a model-agnostic and \textbf{P}rimitive-\textbf{B}ased \textbf{Adv}ersarial training (PBadv) method to deal with this problem. Besides, the latest studies highlight the weakness of the perception of hard compositions even under data-balanced conditions. To this end, we propose a novel over-sampling strategy with object-similarity guidance to augment target compositional training data. We performed detailed quantitative analysis and retrieval experiments on well-established datasets, such as UT-Zappos50K, MIT-States, and C-GQA, to validate the effectiveness of our proposed method, and the state-of-the-art (SOTA) performance demonstrates the superiority of our approach. The code is available at \textcolor{red}{\url{https://github.com/lisuyi/PBadv_czsl}}.
\end{abstract}



\received{20 February 2007}
\received[revised]{12 March 2009}
\received[accepted]{5 June 2009}

\maketitle

\section{Introduction}
Humans can easily perceive novel visual compositions (or pairs) of attributes and objects. For example, if someone has seen a \textit{green apple} and a \textit{sliced strawberry}, they would immediately recognize a \textit{green strawberry}, even if they have never encountered this special composition before. This phenomenon arises from the human ability to proficiently decompose known visual entities into more basic conceptual primitives and subsequently compose them in various compositions~\cite{lake2014towards}. To endow this fundamental ability to vision models, researchers~\cite{2017_red_wine,2018_attributes_operators} have proposed the task of Compositional Zero-shot Learning (CZSL), which aims to efficiently generalize known visual concepts from seen compositions to novel unseen compositions. 

As highlighted in recent outstanding works~\cite{2021_cge,2022OADis,2023_learning_conditional,2023_cot}, the primary challenge of the CZSL task stems from the complex contextual relationship between attributes and objects, resulting in the attribute's significant visual variations across different objects. To address this issue, A basic disentangling strategy~\cite{2022OADis,2022_decomposable_causal} has been proposed. Nonetheless, such a straightforward disentangling approach is confronted with a clear limitation: The oversimplified separation modules, combined with the inherent issue of data imbalance~\cite{2023_cot}, result in the learned visual primitives being overly fragile. Meanwhile, the latest state-of-the-art studies~\cite{2023_learning_conditional,2023_cot} emphasized using object information as guidance, overlooking exploring the mechanisms of contextual interactions between visual primitives.

\begin{figure}[!t]
\small
\centering
  \includegraphics[width=0.6\linewidth]{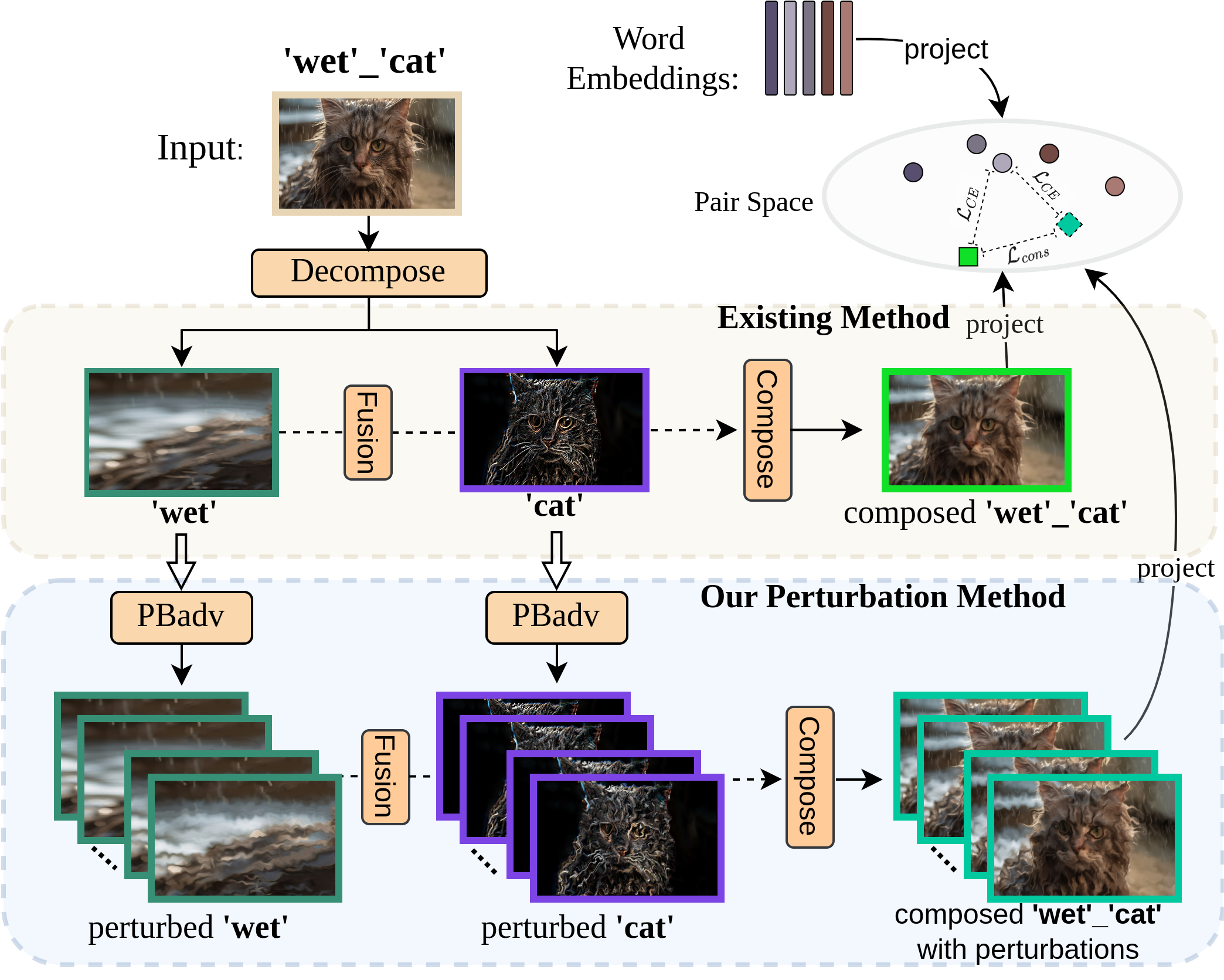}
  \caption{Motivation illustration. Previous state-of-the-art works (denoted by a yellow dashed border) lacked further exploration into the semantic interaction mechanism in the visual domain. To this end, we devise a perturbation method (denoted by a blue dashed border) named PBadv to model the mechanism of the complex interactions to allow the CZSL model to be robust to visually diverse compositions.}
    \label{fig:motivation}
\end{figure}

Inspired by the success of adversarial training methods~\cite{2014_fgsm, 2017_pgd,2022_adv_moment} in Cross-Domain Few-Shot Learning~\cite{2023_styleadv,cm_adv_1,cm_adv_2}, we propose a novel and model-agnostic \textbf{P}rimitive-\textbf{B}ased \textbf{Adv}ersarial training method (PBadv) for CZSL (as shown in Fig. \ref{fig:motivation}). PBadv is devised to alleviate the vulnerability of visual primitives by establishing the contextual semantic interaction mechanism as a random perturbation process, coupled with the perturbation of Gaussian noise. Specifically, it executes a one-step gradient-based attack on the decomposed state and object visual primitives. This operation derives gradients from the semantic embedding space in the opposite direction, subsequently generating $virtual$ perturbed features. A range of synthesized $virtual$ features with diverse perturbation levels can be regarded as simulations of the variation in visual primitives under contextual interaction. The perturbed $virtual$ features are considered to be more challenging for the network to identify, thus increasing the loss of the network. Meanwhile, PBadv optimizes the overall model by minimizing the losses of classification accuracy with both $true$ and $virtual$ features. The minimax optimizing strategy simulates the complex interactions in compositional features while effectively improving the robustness of the model to variable primitives. 

Additionally, previous studies~\cite{2023_learning_conditional,2023_cot} have highlighted the hubness problem caused by the long-tailed distribution in real-life datasets. The severe data-imbalanced problem leads to indistinguishable aggravated visual features in embedding spaces, which restrict the model's generalization ability in CZSl tasks. Besides, recent studies~\cite{2024_jiang} have revealed an intriguing phenomenon: classification performance remains relatively poor towards specific compositions even under conditions of data balance. Both of these phenomena indicate that the model is under-fitting towards certain compositions in the training phase. Inspired by previous outstanding studies~\cite{2022OADis}, we re-purpose the disentangling network to synthesize desired compositional features for data augmentation.
Additionally, we have observed an intriguing phenomenon: similar objects often share common attributes, and their visual features tend to be alike. To take this prior knowledge into our data augmentation approach, we propose a novel \textbf{O}bject-\textbf{S}imilarity-Based \textbf{O}ver\textbf{S}am\textbf{p}ling Method (OS-OSP) to augment specific compositional classes. We avoid directly sampling from the original training dataset, in contrast, we select other semi-positive compositional samples to form corresponding attribute-object pairs via the above disentangling architecture. Notably, the object-similarity-based method utilizes semantic information from language models (\eg, GloVe~\cite{2014glove}), significantly improving the quality of the synthesized compositional features. 

Our contributions can be summarized as follows:
\begin{itemize}
    \renewcommand\labelitemi{\textcolor{black}{$\bullet$}}
    \item We propose a novel and model-agnostic primitive-based adversarial training method (PBadv) for CZSL. PBadv treats the contextual interactions as a random perturbing process, improving the model's robustness to visual variable compositions and the generalization ability towards unseen targets.
    \item We have developed an Object-Similarity-Based Over-Sampling Method (OS-OSP), which aims to alleviate the problem of indistinguishable aggravated embedded visual features and enhance the model's performance on challenging compositions.

    \item We achieve consistent SOTA performance on three well-established datasets and conduct extensive experiments, including classification and retrieval, to evaluate the effectiveness of our method.
\end{itemize}

\section{Related Work}
\subsection{Compositional Zero-shot Learning}
Compositional Zero-shot Learning (CZSL) is tasked to identify novel combinations of attributes and objects by learning visual compositionality from the training phase. Early research in this field can be broadly divided into two primary streams: one~\cite{2017_red_wine,2020_symnet,2021_revisiting_visual_product,2022_kgsp} endeavors to convert the predictions for compositions into individual predictions for objects and states, while the others \cite{2020_hierarchical_decompostition,2021_cge,2021_ow_czsl} regard compositions as unified entities and apply a disentangling strategy. However, separately predicting makes the network susceptible to background noise interference and impedes the network's ability to focus on the correct compositions effectively. To this end, Yang \etal ~\cite{2020_hierarchical_decompostition} proposed a hierarchical decomposing and composing architecture, which predicts compositions and corresponding attribute and object primitives in three hierarchical forms. OADis~\cite{2022OADis} utilized decomposed features to hallucinate virtual features to represent the seen and unseen compositions to regularize the model learning better. Jiang~\etal treated the CZSL task as
a proximate class imbalance problem and employed the logit adjustment method to calibrate the posterior probability for individual classes. Besides, some studies constrain the word vectors of compositions. For instance, the graph convolution method~\cite{2021_cge} has been proposed to propagate semantic connections between compositions in linguistic space, while CANet~\cite{2023_learning_conditional} integrated object word vector information into attribute word vectors, narrowing down the selection space for attributes in the model. These studies treat compositions as a unified entity, and the implementation of decoupling and coupling mechanisms can significantly improve the model's semantic comprehension and its capability to generalize unseen targets.

With the widespread adoption of large-scale models like Vision Transformer (ViT-B)~\cite{2021_vit} and CLIP~\cite{clip} in the field of computer vision, some researchers have successfully replaced traditional backbones (\eg, ResNet18~\cite{2016_resnet}) applied in CZSL tasks and achieved significant performance improvements. Hao~\etal~\cite{2023learning_attention_disentangler} first applied Vision Transformer as a replacement for the traditional ResNet18 backbone, allowing the model to capture long-range dependencies better. Similarly, CoT~\cite{2023_cot} devises a hierarchical structure comprising object and attribute experts to generate compositional features, fully leveraging the capabilities of the entire visual backbone. Additionally, previous outstanding works~\cite{2023learning_composed_soft,2023decomposed_soft,2024caila} based on CLIP have also achieved remarkable results. These arts~\cite{2023learning_composed_soft,2023decomposed_soft,2024_clip_gipcol} primarily focus on constructing more flexible textual prompts. For example, Nihal \etal~\cite{2023learning_composed_soft} proposed to compose prompts as soft learnable tokens, allowing the prompts to be tuned to recognize multiple classes. In this study, we primarily focus on the impact of contextual interactions on the visual primitives of attributes and objects. By rethinking complex semantic interactions as a random perturbation process, our method improves the robustness of the network toward visual variable compositions. 

\subsection{Adversarial Attack}
The target of adversarial attacks is to mislead the classifier by generating an example in the vicinity of the input data. The existing methods for generating adversarial examples can be summarized into three groups~\cite{2018boost_adv}. One of the most representative works is the fast gradient sign method (FGSM~\cite{2014_fgsm}), which generates adversarial examples by incorporating the sign of the gradient of the loss function (\eg, the cross-entropy loss) into the original image. This process is achieved by maximizing the loss function and is widely recognized as a one-step gradient-based method. In parallel, the \textbf{iterative methods}, such as PGD~\cite{2017_pgd}, improve the process of the one-step method by performing multiple iterative attacks instead of a single attack. The iterative methods are proven to better white-box adversaries compared to one-step methods~\cite{2017_ensemble_adv,2018boost_adv}. Besides, Szegedy~\etal~\cite{2013intriguing} proposed a \textbf{optimization-based} method, which utilizes a box-constrained L-BFGS to approximate the chosen process to find a minimizer to make the generated adversarial example to be correctly classified. Adversarial attacks have been widely applied to improve or evaluate the robustness of deep neural networks~\cite{2016_adv_def,2017uni_adv,2016delv_adv}. 

Currently, adversarial attacks have been extensively utilized in downstream applications. Wang~\etal~\cite{2024_adv-ap1} proposed a multi-task adversarial attack algorithm and improved the robustness of deep-learning-based identity management systems towards various face datasets. To deal with the problem of input data agnostic, Duan~\etal~\cite{2022_adv-gen} devise a multi-sample generation model to learn the distribution of the original dataset followed by identical sub-networks to generate the corresponding targeted samples. A novel space-mapping attack method has been introduced into the field of unsupervised subspace
clustering~\cite{2023a2sc} and illustrates the outstanding performance of cross-task transferability. Recently, several studies attempted to explore adversarial learning in Few-shot Learning (FSL), MetaAdv~\cite{2021meta_adv} first proposed to attack the input pictures, significantly improving the defense ability against adversarial targets. StyleAdv~\cite{2023_styleadv} attempted to deal with the domain shift problem by progressively attacking multiple feature spaces. In this study, we develop a one-step attack strategy at the feature level, simulating the variation in visual appearance induced by contextual interaction in CZSL tasks.


\section{Our Approach} \label{sec:approach}

\begin{figure*}[!t]
\small
\centering
  \includegraphics[width=1.0\linewidth]{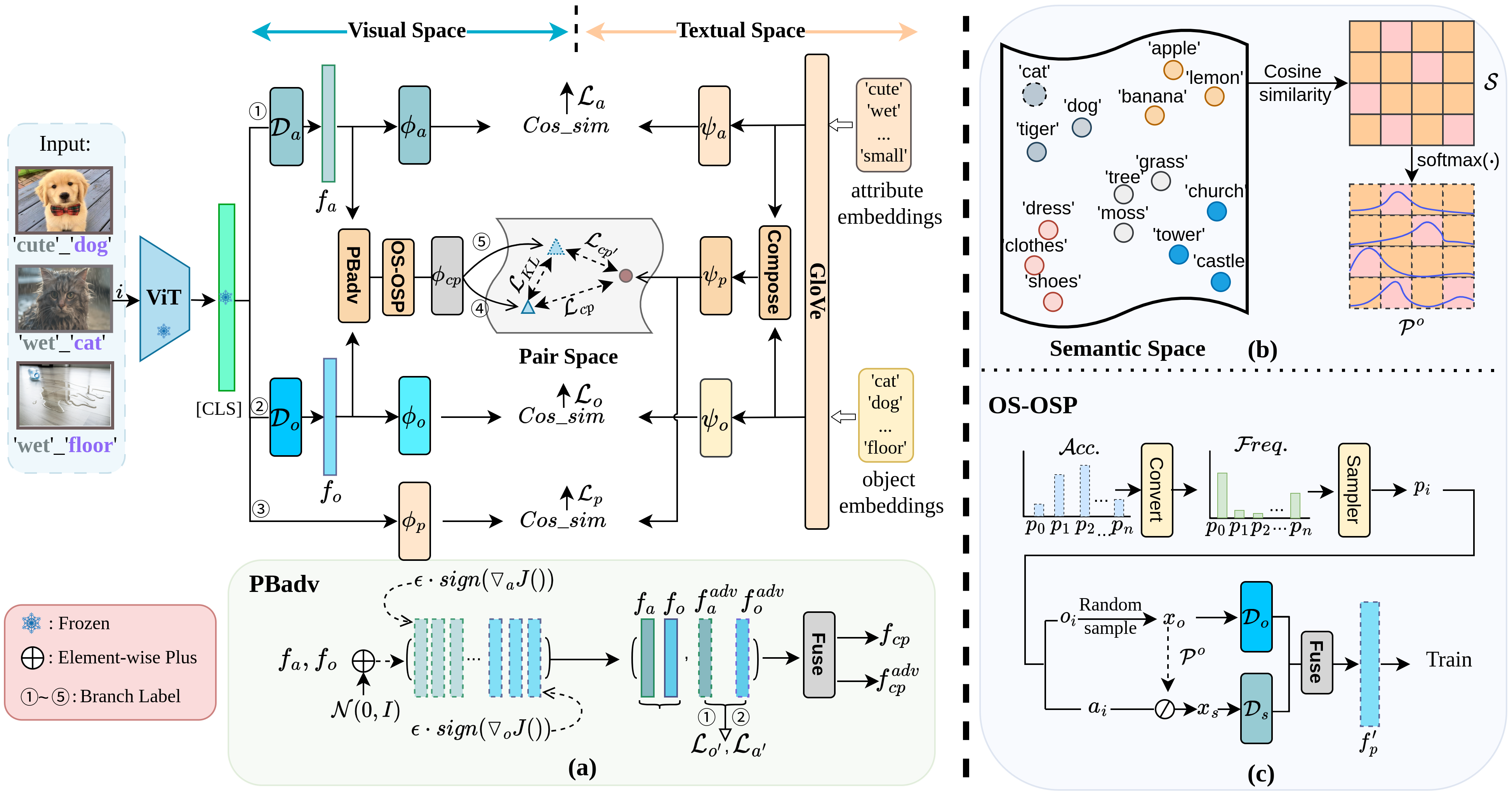}
    \caption{An Overview of PBadv. The framework of the classification network can be divided into two parallel parts: the visual primitive adversarial training branch (\ding{192}, \ding{193}, \ding{195} and \ding{196}) and the base branch (\ding{194}). Branch \ding{192} and \ding{193} are traditional disentangling architecture, which aims to decompose encoded visual features (\ie, [CLS]) into corresponding attribute and object features (\ie, $\boldsymbol{f}_a$ and $\boldsymbol{f}_o$). Next, a series of perturbed features will be synthesized by adding Gaussian noise and attacking the original features (our PBadv method, shown as (a)). After the compositional feature rebuilding process (\ding{195} and \ding{196}), the composed original and perturbed features will be projected into a common pair space for semantic alignment. (b): The object-similarity-guidance sampling probability $\mathcal{P}^{o}$ can be achieved by computing the similarity between all possible objects via prior knowledge (\eg, GloVe). (c): In the OS-OSP method, the probability $\mathcal{P}^{o}$ will be referenced to guide the attribute selection of images during the oversampling process.}
    \label{fig:main-arch}
\end{figure*}

In this section, a detailed description of our method is provided. Initially, the formal definition of the CZSL task is presented, followed by the necessary preliminaries. Subsequently, the comprehensive implementation of our novel PBadv and OS-OSP data augmentation methods is introduced. The overall architecture is depicted in Fig. \ref{fig:main-arch}.

\subsection{Task Formulation and Preliminary}
Unlike traditional classification tasks, CZSL predicts the target as a composition (or pair) of attribute and object (\ie, $y$=$(a,o)$). If we denote the set of states as $\mathcal{A}=\{a_0,a_1,a_2,...,a_m\}$ and the set of objects as $\mathcal{O}=\{o_0,o_1,o_2,...,o_n\}$, thus the set of all possible compositions $\mathcal{Y}$ can be represented as $\mathcal{Y}=\mathcal{A}\times\mathcal{O}$, and the total number of compositions can be calculated as $|\mathcal{Y}|=m\cdot n$. Specifically, in CZSL settings, the set $\mathcal{Y}$ should be divided into two disjoint parts, the set of seen compositions $\mathcal{Y}_{seen}$ and the unseen set $\mathcal{Y}_{unseen}$. In the generalized closed-world scenario, the CZSL model is required to classify both seen and unseen compositions (\ie, $\mathcal{Y}_{test} = \mathcal{Y}_{seen} \cup \mathcal{Y}_{unseen}$), with the constrict that the sum of the cardinalities of the seen and unseen sets is less than that of the total set (\ie, $|\mathcal{Y}_{seen}|+|\mathcal{Y}_{unseen}|<|\mathcal{Y}|$). In the open-world setting, the test examples can be any of the possible compositions, \ie, $\mathcal{Y}_{test}$ = $\mathcal{Y}$. Thus, the task of CZSL can be formulated into $f= \mathcal{X} \rightarrow \mathcal{P}_{test}$, where $\mathcal{X}$ is the set of input RGB image $x$.

\noindent\textbf{Semantic Embedding Space.} The mainstream approach involves projecting the visual features of input images and auxiliary word vectors into a semantic embedding space for alignment. Typically, cosine similarity~\cite{2021_cge,2022OADis,2023learning_attention_disentangler} is widely applied to estimate the compatibility score in common space. 

Denote the embedded visual features as $\boldsymbol{v}$ and the embedded word vectors as $\boldsymbol{w}$, the compatibility score in the embedding space can be formulated as follows:
\begin{equation}\label{equ:equ1}
\mathcal{C} = Cos(\boldsymbol{v},\boldsymbol{w}) = \frac{\boldsymbol{v}^{T}\cdot \boldsymbol{w}}{\|\boldsymbol{v}\|_2\cdot \|\boldsymbol{w}\|_2}.
\end{equation}

For example, given the extracted attribute feature $\boldsymbol{f}_a$, the compatibility score for a specific attribute $\boldsymbol{w}_{a_i}$ can be computed as:
\begin{equation}\label{equ:equ2}
\mathcal{C}_a(\boldsymbol{f}_a,\boldsymbol{w}_{a_i}) = \frac{\phi(\boldsymbol{f}_a)\cdot \psi(\boldsymbol{w}_{a_i})}{\|\phi(\boldsymbol{f}_a)\|_2\cdot \|\psi(\boldsymbol{w}_{a_i})\|_2},
\end{equation}
where $\mathcal{C}_a$ denotes the score function for the attributes, and the corresponding visual and textual projection functions are $\phi$ and $\psi$. The compatibility score for the objects (\ie, $\mathcal{C}_o$) and the compositional pairs (\ie, $\mathcal{C}_p$ and $\mathcal{C}_{cp}$) can be calculated similarly. 

Additionally, we notice that the Ensemble Learning method~\cite{2018_ensemble} is widely employed as a technique to improve the classification performance in recent CZSL works~\cite{2023_learning_conditional,2023learning_attention_disentangler,2023_cot,2020_hierarchical_decompostition}. In this approach, the final prediction is derived by aggregating the predictions from various branches, treating them as votes with different weights during the inference phase. Similar to previous studies, we employ the ensemble learning approach for inference in this paper.

\subsection{Primitive-Based Adversarial Training}
The core problem in CZSL tasks is the significant visual discrepancies caused by the intricate contextual interactions between attributes and objects. To the best of our knowledge, none of the existing works have explored the mechanism of visual contextual interactions from the point of adversarial training~\cite{2014_fgsm,2017_pgd}. We observe that the alteration of associated visual primitives induced by contextual interaction can be construed as a random perturbation through adversarial attacks on the visual primitives. Before delving into our method (\ie, PBadv), we provide a brief overview of the prevailing disentangling strategy~\cite{2022_decomposable_causal,2022OADis, 2024_jiang,2023learning_attention_disentangler}.

\noindent\textbf{Decomposing and Composing Architecture.}  To decompose the entangled visual compositions into primitives of attributes and objects, the simple and direct MLP-based~\cite{2022OADis,2022_decomposable_causal,2023_cot} or Attention-based~\cite{2023learning_attention_disentangler} methods have been widely adopted. The first stage targets disentangling compositional visual features into corresponding visual primitives of attributes and objects (\ie, branches denoted by \ding{192} and \ding{193} in Fig. \ref{fig:main-arch}). For simplicity, we denote the decomposing function as $\mathcal{D}_a$ and $\mathcal{D}_o$ respectively, thus the decomposed visual primitives for attribute $\boldsymbol{f}_a$ and object $\boldsymbol{f}_o$ can be achieved by:
\begin{equation}\label{equ:equ3}
\boldsymbol{f}_a = \mathcal{D}_a(\boldsymbol{f}_{cls}),\ \boldsymbol{f}_o = \mathcal{D}_o(\boldsymbol{f}_{cls}),
\end{equation}
where the class token $\boldsymbol{f}_{cls}$ represent extracted visual feature from the backbone (\ie, ViT-B). Cross-entropy losses are separately incorporated to optimize the modules and force the decomposing function to disentangle the correct components:

\begin{equation}\label{equ:equ4}
\mathcal{L}_a = -\sum_{a\in \mathcal{A}}^{}\log_{}{\frac{exp^{(\frac{1}{\tau} \cdot \mathcal{C}_{a}( \boldsymbol{f}_a,\boldsymbol{w}_a))} }{ {\textstyle \sum_{a'\in \mathcal{A}}^{}exp^{(\frac{1}{\tau} \cdot \mathcal{C}_{a}(\boldsymbol{f}_a,\boldsymbol{w}_{a'})) }}  }},
\end{equation}

\begin{equation}\label{equ:equ5}
\mathcal{L}_o = -\sum_{o\in \mathcal{A}}^{}\log_{}{\frac{exp^{(\frac{1}{\tau} \cdot \mathcal{C}_{o}( \boldsymbol{f}_o,\boldsymbol{w}_o))} }{ {\textstyle \sum_{o'\in \mathcal{O}}^{}exp^{(\frac{1}{\tau} \cdot \mathcal{C}_{o}(\boldsymbol{f}_o,\boldsymbol{w}_{o'}))}}  } } ,
\end{equation} where $\tau$ is temperature coefficient~\cite{2019_cos-tmp}. 

The subsequent stage entails the reconstitution of the disentangled attribute and object primitives into the corresponding pair feature $\boldsymbol{f}_{cp}$ (Noting: for clarity and consistency with previous studies, we use the notation $\boldsymbol{f}_{cp}$ or $\boldsymbol{f}_{p}$ instead of $\boldsymbol{f}_{cy}$ or $\boldsymbol{f}_{y}$). Consistent with prevailing research, we utilize a \textbf{Fusion} function (as depicted in Fig. \ref{fig:main-arch}), which involves concatenating attribute and object primitive features and projecting them through an MLP function $\phi_{cp}$. The optimization process is almost identical to the above decomposing part. Thus, we can compute the loss $\mathcal{L}_{cp}$ for the composed compositional features:
\begin{equation}\label{equ:equ6}
\mathcal{L}_{cp} = -\sum_{y\in \mathcal{Y}_{train}}^{}\log_{}{\frac{exp^{(\frac{1}{\tau} \cdot \mathcal{C}_{p}( \boldsymbol{f}_{cp},\boldsymbol{w}_y))} }{ {\textstyle \sum_{y'\in \mathcal{Y}_{train}}^{}exp^{(\frac{1}{\tau} \cdot \mathcal{C}_{p}(\boldsymbol{f}_{cp},\boldsymbol{w}_{y'}))}}  }}.
\end{equation}

Notably, the compositional word embedding 
$ \boldsymbol{w}_y$ is achieved by a composing function (as illustrated in Fig. \ref{fig:main-arch}), and its implementation is symmetrically consistent with the visual composing part. 

Despite the decoupling architecture enhancing the model's semantic comprehension, treating attribute-object pairs as an integrated entity remains a fundamental branch for its superior robustness against background noise in images. Thus, we retain this base branch in our method (\ie, branch \ding{194} in Fig. \ref{fig:main-arch}) and incorporate its loss $\mathcal{L}_p$ as follows:
\begin{equation}\label{equ:equ7}
\mathcal{L}_{p} = -\sum_{y\in \mathcal{Y}_{train}}^{}\log_{}{\frac{exp^{(\frac{1}{\tau} \cdot \mathcal{C}_{p}( \boldsymbol{f}_{cls},\boldsymbol{w}_y))} }{ {\textstyle \sum_{y'\in \mathcal{Y}_{train}}^{}exp^{(\frac{1}{\tau} \cdot \mathcal{C}_{p}(\boldsymbol{f}_{cls},\boldsymbol{w}_{y'}))}}  }}.
\end{equation}

It is worth noting that our primary contribution lies in rethinking the problem of semantic interaction in CZSL as a process of random and adversarial attacking manner, we choose to use the class token of ViT-B (\ie,$\boldsymbol{f}_{cls}$) as our visual input for the classification network in our method rather than utilize more intra-layer information~\cite{2022OADis,2023_cot,2024_clip_gipcol}.

\noindent\textbf{Primitive-Level Perturbation Algorithm and Consistency Loss.}
Inspired by the famous attacking methods FGSM~\cite{2014_fgsm} and the successful practice of feature-level attacking method in the field of Cross-Domain Learning~\cite{2023_styleadv,cm_adv_1}, we propose to model the process of visual prototype variation resulting from semantic interaction as a random adversarial attacking procedure. Since the visual primitives disentangling and recomposing are directly conducted in the visual semantic domain, we do not attack the target at the pixel level but employ a feature level~\cite{2023_styleadv} and one-step attack method to generate the perturbed adversarial visual primitives $\boldsymbol{f}^{adv}$:
\begin{equation}\label{equ:equ8}
\boldsymbol{f}^{adv}_a = \boldsymbol{f}_a +  \epsilon \cdot sign(\triangledown_{\boldsymbol{f}_a} J(\theta_a, \boldsymbol{f}_a, a)),
\end{equation}
\begin{equation}\label{equ:equ9}
\boldsymbol{f}^{adv}_o = \boldsymbol{f}_o + \epsilon \cdot sign(\triangledown_{\boldsymbol{f}_o} J(\theta_o, \boldsymbol{f}_o, o)),
\end{equation}
where $J(\cdot)$ is the object function, $\theta_a$ and $\theta_o$ are learnable parameters of classification modules for attribute and object respectively. The signed gradients of the $\boldsymbol{f}_a$, $\boldsymbol{f}_o$ can be achieved from the backward propagation chain by computing the corresponding classification loss. Besides, to guarantee that the disturbed primitives preserve consistent class information as the original prototypes, we compute $\mathcal{L}^{adv}_{a}$ and $\mathcal{L}^{adv}_{o}$ by applying the identical cross-entropy loss as illustrated in Eqs. (\ref{equ:equ4}) and (\ref{equ:equ5}).

Although we endeavor to employ the adapted FGSM method to mimic the inherent contextual interactions characteristic of CZSL tasks, we aspire to endow the generated adversarial features with greater diversity. To this end, we do not fix the hyper-parameter $\epsilon$ as a constant. Instead, $\epsilon$  will be randomly selected from a manually constructed list $l_{\epsilon}$ to generate diverse perturbation levels as a simulation of the visual variation under contextual interaction, which is consistent with the outstanding work StyleAdv~\cite{2023_styleadv} but with different purposes. Besides, we incorporate the random Gaussian noise into the feature primitives before applying the FGSM algorithm:
\begin{equation}\label{equ:equ10}
\boldsymbol{f'}_a = \boldsymbol{f}_a + k\cdot \mathcal{N}(0,I),\ \boldsymbol{f'}_o = \boldsymbol{f}_o + k\cdot \mathcal{N}(0,I),
\end{equation}
where $k$ is a weighting coefficient used to control the ratio of noise influence.
This operation facilitates PBadv in generating adversarial feature primitives with increased randomness, thus making the attacking process closer to real contextual interactions in CZSL tasks. It helps the synthesized perturbed features exhibit more diversity rather than aiming for a better random start under the iterative setting~\cite{2017_pgd}. 

We concurrently derive the perturbed composed feature $\boldsymbol{f}^{adv}_{cp}$ synthesized from the perturbed attribute and object primitives, alongside the original composed features $\boldsymbol{f}_{cp}$. It's important to note that we do not generate perturbed features $\boldsymbol{f}^{adv}_{cp}$ by directly attacking the original composed features $\boldsymbol{f}_{cp}$ in embedding space. Instead, we treat the compositional features synthesized from the perturbed primitives as $\boldsymbol{f}^{adv}_{cp}$, considering that the ultimate goal of our PBadv is to simulate the contextual interactions and enhance the robustness of existing disentangling networks to the variable feature prototypes. The other question is semantic information retaining. On the one hand, we can utilize the above-mentioned cross-entropy loss and compute corresponding $\mathcal{L}_{cp}$ and $\mathcal{L}^{adv}_{cp}$. On the other hand, we can design similarity metric losses from the perspective of feature distribution obtained from embedding functions. To this end, a straightforward and intuitive approach is to compute the Kullback-Leibler (KL) divergence, thus we compute the KL loss $\mathcal{L}_{KL}$ by:
\begin{equation}\label{equ:equ11}
\mathcal{L}_{KL} = KL(\phi_{cp}(\boldsymbol{f}_{cp}),\phi_{cp}(\boldsymbol{f}^{adv}_{cp})),
\end{equation}
thus, the consistency loss between the $\boldsymbol{f}_{cp}$ and $\boldsymbol{f}^{adv}_{cp}$ can be summarized as:
\begin{equation}\label{equ:equ12}
\mathcal{L}_{cons} = \mathcal{L}^{adv}_{cp} + KL(\phi_{cp}(\boldsymbol{f}_{cp}),\phi_{cp}(\boldsymbol{f}^{adv}_{cp})).
\end{equation}

\subsection{Object Similarity Based Oversampling Method}

Although our novel PBadv method significantly enhances the model's robustness to visually variant prototypes, the inherent data long-tailed distribution and severe class imbalance noticeably constrain the performance of the classification network~\cite{2023learning_attention_disentangler,2023_cot,2024_jiang}. Furthermore, Jiang \etal \cite{2024_jiang} have unveiled the presence of challenging compositions that lead to performance degradation, even in the presence of balanced training data. To address this, a straightforward approach is to augment the minority samples in the training set~\cite{2023_cot}. In contrast, we treat the minority or hard compositions as under-fitting targets and introduce an adaptive oversampling strategy to deal with this issue during the training phase (illustrated in (c) of Fig. \ref{fig:main-arch}). Before each epoch of model optimization, we compute the goodness-of-fit of each compositional class $y_k$ ($y_k \in \mathcal{Y}_{train}$) by comparing the predicted results with the true labels, denoted as $\mathcal{A}cc(y_k)$. Next, we can convert $\mathcal{A}cc(y_k)$ and achieve the over-sampling frequency by applying the soft-max function():
\begin{equation}\label{equ:equ13}
\mathcal{F}req(y_k) = \frac{exp^{(\alpha-\mathcal{A}cc(y))}}{\sum_{y_i\in \mathcal{Y}_{seen}}^{}exp^{(\alpha-\mathcal{A}cc(y_i))}},
\end{equation} 
where $\alpha$ is the hyper-parameter to control the sampling frequency. Then, the \textbf{Sampler} (as shown in (c) of Fig.  \ref{fig:main-arch}) will sample the corresponding positional classes. In this study, we set $\alpha = 0$ for simplicity.
Subsequently, the decomposing and composing architecture can be re-purposed to synthesize virtual compositional features to augment the under-fitting targets. 

Furthermore, we have observed an interesting phenomenon: similar objects often share common attributes, and their attribute visual features tend to be alike.
For instance, if there is a $cute\ dog$, there may also be a $cute\ cat$ in reality, as cats and dogs are similar entities.
Additionally, compared to a $wet\ floor$, the visual features of a $wet\ cat$ and a $wet\ dog$ are more similar, as their fur tends to curl rather than darken in color. This inspires us to incorporate such prior information into our over-sampling algorithm, allowing the synthesized virtual compositional features to be closer to reality. Firstly, we treat the pre-trained linguistic model (e.g., GloVe) as an external knowledge base and transform the semantic similarity between objects into cosine similarity between corresponding word embeddings (sub-figure (b) of Fig. \ref{fig:main-arch}). Thus, the similarity map among objects can be achieved by:
\begin{equation}\label{equ:equ14}
\mathcal{S} = [\frac{w_0}{||w_0||} : \frac{w_{n-1}}{||w_{n-1}||}]^{T} \times [\frac{w_0}{||w_0||} : \frac{w_{n-1}}{||w_{n-1}||}].
\end{equation} 

Next, we utilize the softmax function to transform the similarity between objects into the sampling frequency of objects $o_i$, denoted as $\mathcal{P}(o_i)$:
\begin{equation}\label{equ:equ15}
\mathcal{P}(o_i) = \frac{exp^{(\mathcal{S}_{(i,0)})}}{\sum_{j:0 \sim n-1}^{}exp^{(\mathcal{S}_{(i,j)})}}.
\end{equation}

Notably, we argue that this approach yields features with greater diversity than directly oversampling from the original dataset, and we denote this method as the Object-Similarity-Based OverSampling (OS-OSP) method. In this approach, we prioritize sampling images from objects with semantic similarity when synthesizing compositional features for the target class. This allows the recomposing network to achieve suitable attribute features and ensures that the synthesized virtual features exhibit greater diversity and adhere close to reality. Furthermore, we design OS-OSP as a plug-and-play module, which can be seamlessly integrated into the training process. 

\subsection{Training and Inference}
Incorporation of the conventional disentangling network and base classification branch, our final loss calculation involves $\mathcal{L}_a$, $\mathcal{L}_o$,  $\mathcal{L}_p$ and $\mathcal{L}_{cp}$. For simplistic, we define the loss for the conventional classification network as $\mathcal{L}_{base}$:
\begin{equation}\label{equ:equ16}
\mathcal{L}_{base} = \mathcal{L}_p + \underbrace{\mathcal{L}_{a} + \mathcal{L}_{o} + \mathcal{L}_{cp}}_{\text{Disentangling branch}}.
\end{equation} 

In addition, our PBadv algorithm synthesized a series of adversarial perturbed features. To maintain semantic consistency, we incorporate the $\mathcal{L}^{adv}_{a}$, $\mathcal{L}^{adv}_{o}$ and $\mathcal{L}_{cons}$ into the final loss, and denoted as $\mathcal{L}_{adv}$:
\begin{equation}\label{equ:equ17}
\mathcal{L}_{adv} = \mathcal{L}^{adv}_{a} + \mathcal{L}^{adv}_{o} + \mathcal{L}_{cons}. 
\end{equation} 
Thus, the final training loss linearly combines the two losses above:
\begin{equation}\label{equ:equ18}
\mathcal{L} = \mathcal{L}_{base} + \mathcal{L}_{adv}. 
\end{equation} 

In the inference phase, we apply an assembling strategy and achieve the final compositional compatibility score by considering the classification branches \ding{192}, \ding{193}, \ding{194}, and \ding{195} together:

\begin{equation}\label{equ:equ19}
\mathcal{C}(y=(s,o)) = \mathcal{C}_{p} + \mathcal{C}_{a} + \mathcal{C}_{o} +\mathcal{C}_{cp},
\end{equation}
where, each $\mathcal{C}_{*}$ is calculated with Eq. \ref{equ:equ2}, and the final classification result follows the highest value in Eq. \ref{equ:equ19}.

\section{Experiments}
In this section, we provide a detailed description of the implementation of our model and the experiment settings. Specifically, we will sequentially introduce the datasets used in the experiments, the evaluation metrics, the quantitative experimental results, and the ablation experiments to demonstrate the effectiveness of our method. Finally, we will present a qualitative analysis of the retrieval performance of our model.

\subsection{Experimental Setting}
\begin{table}[!t] 
\small
\centering
\caption{The detailed splits for three widely used datasets: UT-Zappos50K, MIT-States, and C-GQA. The symbols $a$ and $o$ represent the total number of states and objects in each corresponding dataset. Correspondingly, $i$ denotes the total number of images. $sp$ and $up$ represent the number of seen and unseen compositional classes.}
\setlength{\tabcolsep}{6.0 pt}
\begin{tabular}{lcccccccccc}
\toprule
\multirow{2}{*}{Dataset} & & &\multicolumn{2}{c}{\textbf{Training}} &\multicolumn{3}{c}{\textbf{Validation}} &\multicolumn{3}{c}{\textbf{Test}} \\
\cmidrule(lr){4-5} \cmidrule(lr){6-8} \cmidrule(l){9-11}
  &$a$ &$o$  & $sp$ &$i$   &$sp$ &$up$ &$i$   &$sp$ &$up$ &$i$ \\
\midrule
C-GQA~\cite{2021_cge}      &413 &674   &5592 &27k   &1252 &1040 &7k   &888 &923 &5k  \\
MIT-States~\cite{2015_mitstates} &115 &245   &1262 &30k  &300 &300 &10k   &400 &400 &13k \\
UT-Zap.~\cite{2014_utzappos}  &16 &12    &83 &23k   &15 &15 &3k   &18 &18 &3k \\
\bottomrule
\end{tabular}
\label{tab:datasets_splits}
\end{table}
\noindent\textbf{Datasets.}
The UT-Zappos50K~\cite{2014_utzappos} dataset constitutes a fine-grained shoe catalog, characterized by its smaller scale and relatively stable and simple content. Its object categories encompass prevalent shoe types, such as $sandal$, $sneaker$, and $boot$, while its attribute classes primarily pertain to shoe materials, including $canvas$, $leather$, and $wool$. The dataset comprises a total of 50,025 images, with 12 categories of shoe types, 16 categories of material attributes, and 116 labeled compositional categories. In contrast, the pictures in the MIT-States dataset~\cite{2015_mitstates} are relatively closer to real-life scenarios and are of moderate scale. This dataset is automatically crawled and annotated from websites using retrieval algorithms, thus inherently containing noticeable noise. It comprises 1962 labeled compositional categories, including 115 states and 245 objects. The C-GQA dataset~\cite{2021_cge}, first proposed by Ferjad Naeem et al., presents greater challenges, a larger scale, and clearer annotations. It encompasses a wide array of common objects and states encountered in daily life, including 413 attributes and 674 objects, alongside more than 9,500 compositional labels. 


\noindent\textbf{Evaluation Metrics.}
Consistent with Zero-shot Learning (ZSL), an inherent visual bias exists between seen and unseen compositions in CZSL tasks. Following the approach of previous notable studies~\cite{2021_cge}, we incorporate a finite variable scalar term into the final predictions to achieve optimal results for both seen and unseen compositions. It is evident that there exists an optimal scalar value that can balance the accuracy of both seen and unseen compositions simultaneously. Therefore, we report the best Harmonic Mean (HM), which directly reflects the model's classification capability for both seen and unseen compositions. Additionally, we compute the Area Under the Curve (AUC) across different operational points of the variable scalar. AUC has been widely regarded as a critical metric for assessing the overall classification performance of the model~\cite{2023learning_attention_disentangler,2023_cot} in CZSl tasks. Furthermore, we report the best attribute and object accuracy for unseen targets during the test phase to demonstrate the superiority of our proposed PBadv method.

\definecolor{hm}{RGB}{234, 242, 248} 
\definecolor{auc}{RGB}{224, 250, 240 }
\begin{table*}[!t]
\small
\centering
\caption{The closed-world results on the test set of UT-Zappos50K, MIT-States, and C-GQA. We report the well-established metrics: the best AUC, best HM, best-seen accuracy (Seen), best-unseen accuracy (Unseen), best state accuracy (s), and the best object accuracy (o), highlighting the most important metric--best AUC. For a fair comparison, we reproduce the absent results of current SOTA and representative methods via the public open-source codes. The SOTA results are in bold and the second-to-best results are in underlining. }
\resizebox{\textwidth}{!}{
\begin{tabular}{cl>{\columncolor{auc}}cccccc>{\columncolor{auc}}cccccc>{\columncolor{auc}}cccccc}
\toprule
  &\multirow{2}{*}{Method}  &\multicolumn{6}{c}{UT-Zappos50K}   & \multicolumn{6}{c}{MIT-States}  & \multicolumn{6}{c}{C-GQA}   \\
  \cmidrule(lr){3-8} \cmidrule(lr){9-14} \cmidrule(l){15-20}
  &  & AUC  & HM   & Seen  & Unseen  &Attr.  &Obj.  & AUC  & HM   & Seen  & Unseen  &Attr.  &Obj.    & AUC  & HM  & Seen  & Unseen  &Attr.  &Obj.  \\ 
\midrule
\multirow{7}*{\rotatebox{90}{\textbf{Closed-world}}} 
&SymNet~\cite{2020_symnet} & 32.6 &45.6 &60.6 &68.6 &48.2 &77.0     &5.2 &17.3 &31.0 &25.2 &30.2 &32.3    &3.1 &13.5 &30.9 &13.3 &11.4 &34.6 \\
&CompCos~\cite{2021_ow_czsl} &31.8 &48.1 &58.8 &63.8 &45.5 &72.4   &7.2 &21.1 &34.1 &30.2 &33.6 &37.7   &2.9 &12.8 &30.7 &12.2 &10.4 &33.9  \\
&CGE~\cite{2021_cge}  &34.5 &48.5 &61.6 &70.0 &\textbf{50.8} &\underline{77.1}   &7.3 &21.3 &33.5 &30.9 &33.1 &37.4   &3.8 &15.0 &32.3 &14.9 &13.8 &33.2  \\
&IVR~\cite{2022_IVR}  &34.3 &49.2 &61.5 &68.1 &48.4 &74.6    &3.8 &14.9 &23.2 &25.6 &29.1 &33.6    &2.2 &10.9 &27.3 &10.0 &10.3 &\textbf{37.5}\\
&OADis$^{\dagger}$~\cite{2022OADis}  &32.6 &46.9 &60.7 &68.8 &49.3 &76.9   &7.5 &21.9 &34.2 &29.3 &27.2 &31.9   &3.8 &14.7 &33.4 &14.3 &8.9 &36.3\\
&ADE$^{\dagger}$~\cite{2023_learning_conditional}  &35.1 &51.1 &63.0 &64.3 &46.3 &74.0   &7.4 &21.2 &34.2 &28.4 &32.0 &35.4   &\underline{5.2} &\underline{18.0} &\textbf{35.0} &17.7 &16.8 &32.3 \\
&COT$^{\dagger}$~\cite{2023_cot}  &34.8 &48.7 &60.8 &64.9 &47.3 &73.2   &\underline{7.8} &\underline{23.2} &\underline{34.8} &\underline{31.5} &34.2 &\textbf{40.1}   &5.1 &17.5 &34.0 &\underline{18.8} &14.6 &34.8 \\
\midrule
&\textbf{PBadv}(Ours) &\underline{39.6} &\underline{55.2} &\underline{64.3} &\underline{70.4} &49.2 &76.1  
&7.6  &20.5 &33.5 &28.9 &\underline{32.0} &37.1 
   &4.9 &17.8 &\underline{34.1} &17.8 &\textbf{22.4} &35.7 \\  

&+ \textbf{OS-OSP} &\textbf{40.0} &\textbf{55.6} &\textbf{65.8} &\textbf{71.8} &\underline{50.4} &\textbf{78.8} &\textbf{8.6} &\textbf{23.4} &\textbf{37.6} &\textbf{31.6} &\textbf{35.1} &\underline{39.4}   &\textbf{5.3} &\textbf{18.2} &34.0 &\textbf{18.9} &\underline{18.2} &\underline{37.3} \\  
\bottomrule
\end{tabular}}
\label{tab:cw_results}
\raggedright
\footnotesize The symbol $^{\dagger}$ indicates that the method utilizes intermediate features from the feature extractor.
\end{table*}

\noindent\textbf{Experimental Details.}
We follow the approach of the previous study ADE~\cite{2023learning_attention_disentangler} and utilize the frozen backbone ViT-B-16~\cite{2021_vit} (ViT-B), pre-trained on ImageNet~\cite{2009_imagenet} in a self-supervised manner~\cite{2021_dino}, as the visual feature extractor for our model. Specifically, we directly use the class token ([CLS]), which has a dimension of 768, as the input visual feature. For the conventional disentangling architecture, the implementations for the attribute and object visual disentangling functions, $\mathcal{D}_a$ and $\mathcal{D}_o$, are standard two-layer MLPs with ReLU~\cite{2010_relu}, LayerNorm~\cite{2016_layernorm}, and Dropout~\cite{2014_dropout}. The embedding networks $\phi$ and $\psi$ are structured with a single Fully Connected (FC) layer followed by ReLU. We utilize GloVe~\cite{2014glove} to initialize the 300-dimensional word embedding vectors for attributes and objects. A concatenation operation on the corresponding attribute and object embeddings is applied to achieve the compositional word vectors. In the PBadv block, we Empirically set the hyper-parameter list $\epsilon$ to \{0.0, 0.005, 0.05, 0.5\} and the weighting coefficient $k$ to 16/255. During the training phase, we fix the image backbone and train all other modules using the Adam Optimizer~\cite{2014_adam} with a consistent learning rate $2e-5$ and weight decay $5e-5$ for all three datasets. The countdown of the temperature coefficient $1/\tau$ is set to 40, 20, and 60 for UT-Zappos50K, MIT-States, and C-GQA, respectively. The training process is performed on 2 $\times$ Nvidia GeForce GTX 2080Ti GPUs. The training batch size for all datasets is 128, and all modules in our model are trained for 200 epochs.

\subsection{Quantitative Analysis}
To demonstrate the efficiency of our proposed PBadv method, we compare our experimental results with current state-of-the-art methods, such as ADE~\cite{2023learning_attention_disentangler}, COT~\cite{2023_cot}, and other representative approaches. To ensure fairness, we prioritize reporting the results from their published works. Besides, we re-train the corresponding models using the ViT-B backbone for datasets where results are absent. Followed with ADE~\cite{2023learning_attention_disentangler}, we report the comparative results in both closed and open-world settings, which have been organized into Table \ref{tab:cw_results} and Table \ref{tab:ow_results}.

\definecolor{hm}{RGB}{234, 242, 248} 
\definecolor{auc}{RGB}{234, 250, 241}
\begin{table*}[!t]
\small
\centering
\caption{The open-world results on the test set of UT-Zappos50K, MIT-States, and C-GQA. We report the metrics as mentioned before: the best AUC, best HM, best-seen accuracy (Seen), best-unseen accuracy (Unseen), best state accuracy (s), and the best object accuracy (o).  The most important metric--best AUC is highlighted with color.}
\resizebox{\textwidth}{!}{
\begin{tabular}{cl>{\columncolor{auc}}cccccc>{\columncolor{auc}}cccccc>{\columncolor{auc}}cccccc}
\toprule
  &\multirow{2}{*}{Method}  & \multicolumn{6}{c}{UT-Zappos50K}   & \multicolumn{6}{c}{MIT-States}  & \multicolumn{6}{c}{C-GQA}    \\
  \cmidrule(lr){3-8} \cmidrule(lr){9-14} \cmidrule(l){15-20}
  &  & AUC  & HM   & Seen  & Unseen  &Attr.  &Obj.  & AUC  & HM   & Seen  & Unseen  &Attr.  &Obj.    & AUC  & HM  & Seen  & Unseen  &Attr.  &Obj.  \\ 
\midrule
\multirow{6}*{\rotatebox{90}{\textbf{Open-world}}} 
&SymNet~\cite{2020_symnet} &25.0 &40.6 &60.4 &51.0 &38.2 &\underline{75.0}     &1.73 &8.9 &30.8 &6.3 &21.5 &28.9    &0.77 &4.9 &30.1 &3.2 &18.4 &37.5\\
&CompCos~\cite{2021_ow_czsl} &20.7 &36.0 &58.1 &46.0 &36.4 &71.1   &2.79 &12.0 &33.6 &13.1 &22.9 &33.1   &0.72 &4.3 &32.8 &2.8 &15.1 &37.8  \\
&IVR~\cite{2022_IVR}  &25.3 &42.3 &60.7 &50.0 &38.4 &71.4   &1.60 &9.1 &26.0 &10.3 &20.8 &32.0   &0.94 &5.7 &30.6 &4.0 &16.9 &36.5\\
&OADis$^{\dagger}$~\cite{2022OADis}   &25.3 &41.6 &58.7 &\underline{53.9} &\underline{40.3} &74.7   &2.40 &10.8 &34.7 &10.3 &18.6 &30.9   &0.71 &4.2 &33.0 &2.6 &14.6 &\textbf{39.7}\\
&ADE$^{\dagger}$~\cite{2023_learning_conditional}  &27.1 &44.8 &62.4 &50.7 &39.9 &71.4   &2.51 &11.4 &29.7 &\underline{12.3} &\textbf{22.4} &31.4   &\textbf{1.42} &\textbf{7.6} &\textbf{35.1} &\textbf{4.8} &22.4 &35.6 \\
&COT$^{\dagger}$~\cite{2023_cot}  &25.0 &41.5 &59.7 &50.3 &37.7 &72.1   &\underline{2.97} &12.1 &36.5 &11.2 &20.9 &\textbf{36.4}   &1.02 &5.6 &\underline{34.4} &4.0 &20.7 &35.5 \\
\midrule
&\textbf{PBadv}(Ours) &\underline{29.1} &\underline{45.8} &\underline{66.0} &53.8 &\textbf{48.7} &74.8   &2.86 &\underline{12.1} &\underline{36.8} &11.8 &\underline{23.1} &\underline{33.8}  &{1.06} &{6.1} &{33.3} &\underline{4.1} &\underline{20.7} &37.8 \\

&+ \textbf{OS-OSP} &\textbf{29.7}  &\textbf{46.7} &\textbf{66.5} &\textbf{54.9} &39.5 &\textbf{76.8} 
&\textbf{3.40} &\textbf{13.3} &\textbf{38.0} &\textbf{12.4} &\textbf{23.6} &33.7
&\underline{1.08} &\underline{6.2} &33.7 &4.0     &19.7  &\underline{38.5} \\ 
\bottomrule
\end{tabular}}

\label{tab:ow_results}
\raggedright
\footnotesize The symbol $^{\dagger}$ indicates that the method utilizes intermediate features from the feature extractor.
\end{table*}

\noindent\textbf{Closed-world Evaluation.} We conduct experimental comparisons on the test set with previous outstanding Compositional Zero-shot Learning methods, including SymNet~\cite{2020_symnet}, CompCos~\cite{2021_ow_czsl}, CGE~\cite{2021_cge}, IVR~\cite{2022_IVR}, OADis~\cite{2022OADis}, ADE~\cite{2023learning_attention_disentangler} and COT~\cite{2023_cot}. From Tab. \ref{tab:cw_results}, we observe that the PBadv (ours) method consistently outperforms other works by a large margin and achieves new state-of-the-art results on UT-Zappos50K, MIT-States, and C-GQA. For the most important metric, AUC, we reach the highest values of 40.0\%, 8.6\%, and 5.3\% on three datasets, respectively, with improvements of 4.9\%, 0.8\%, and 0.1\%. Our method achieves new state-of-the-art results for another import metric, improving the harmonic mean (HM) by 4.5\%, 0.2\%, and 0.1\%. The state-of-the-art results on the best-unseen accuracy illustrate our superiority of the OS-OSP augmentation method in improving the generalization ability towards novel targets. Since PBadv aims to enhance the model's robustness towards variable compositional visual appearances, we also report the classification results for attributes and objects. Our method still outperforms other approaches by a significant margin.

\noindent\textbf{Open-world Evaluation.} We also report experimental comparison results in the open-world setting with other methods (as shown in Table \ref{tab:ow_results}). Similarly, PBadv outperforms all other methods on UT-Zappos50K and MIT-States significantly. It is worth noting that we did not replicate the results of the CGE model in the open-world setting. This is because the number of parameters in graph convolution increases geometrically under open-world conditions, making it impractical for real-world applications. We achieve new state-of-the-art AUC results on UT-Zappos50K and MIT-States, with improvements of 2.5\% and 0.43\% compared to other methods. Additionally, we increase the best harmonic mean (HM) from 44.8\%, and 12.1\% to 46.7\%, and 13.3\%, respectively, with improvements of 1.9\%, and 1.2\%. From Table \ref{tab:ow_results}, it is evident that our approach still outperforms prior methods on other metrics by a large margin. The outstanding results in the open-world setting demonstrate the superiority of our proposed PBadv method.

In Table \ref{tab:cw_results} and Table \ref{tab:ow_results}, we also report the best attribute and object accuracy in closed and open-world settings. It is worth mentioning that, unlike ADE~\cite{2023learning_attention_disentangler} and COT~\cite{2023_cot}, we do not use intermediate features from the backbone and fall short in the performance under the dataset of C-GQA. Instead, we designed the classification network to follow the class token directly, ensuring time efficiency in the training phase. Besides, we devise the OS-OSP method into a separable block, facilitating easier model training. Despite using fewer features, the overall performance of PBadv in the metrics of best-seen and best-unseen is comparable to the existing state-of-the-art models or even surpasses them in some datasets. Furthermore, in more critical performance metrics such as AUC, HM, best seen, and best unseen, our model achieves most of the state-of-the-art results, demonstrating the superiority and effectiveness of our proposed PBadv and OS-OSP methods.


\subsection{Effectiveness of PBadv}

\begin{table}[!t]
\small
\centering
\setlength{\tabcolsep}{10.0 pt}
\caption{Comparison results of our method with and without PBadv on the test set of C-GQA~\cite{2021_cge}.}
\begin{tabular}{lcccccc}
\toprule
\multirow{2}{*}{Methods} &\multicolumn{6}{c}{C-GQA}  \\ 
\cmidrule(lr){2-7} 
& \multicolumn{1}{c}{AUC} & \multicolumn{1}{c}{HM} & \multicolumn{1}{c}{Seen} & \multicolumn{1}{c}{Unseen} & \multicolumn{1}{c}{Attr.} & \multicolumn{1}{c}{Obj.} \\
\midrule
\textbf{Base}   &\underline{4.68} &\underline{17.3}  &\textbf{34.2} &\underline{17.0} &\underline{15.9} &\underline{35.2}\\

\textbf{+PBadv} &\textbf{4.89} &\textbf{17.8} &\underline{34.1} &\textbf{17.8} &\textbf{22.4} &\textbf{35.7} \\

\bottomrule
\end{tabular}
\label{tab:abalation_adv}
\end{table}

We first conducted ablation experiments to demonstrate the effectiveness of our proposed PBadv method. Given that the UT-Zappos50K dataset is overly simplistic and the MIT-States dataset contains excessive noise~\cite{2020_mit_noise}, we performed ablation studies (as shown in Table \ref{tab:abalation_adv}) on the more realistic and challenging large-scale C-GQA dataset, which has a greater variety of attributes and object samples.

Specifically, we first tested the basic network, corresponding to the basic branch network described in Eqs (\ref{equ:equ4}), (\ref{equ:equ5}), (\ref{equ:equ6}), and (\ref{equ:equ7}), to evaluate the model's performance on the test set under the closed-world setting. Subsequently, we incorporated our adversarial method into the base model and tested the performance again. As shown in the table, our PBadv method significantly enhances the model's robustness to the variable visual features, substantially improving the overall performance of the model. It boosts AUC and HM from 4.68\% and 17.3\% to 4.89\% and 17.8\%, with improvements of 0.22\% and 0.5\%, respectively. Additionally, we found that PBadv significantly improves the classification accuracy of attributes and objects. We attribute this to the adversarial training method, which substantially enhances the model's robustness to features with large visual variations, thereby improving the classification performance.

\subsection{Ablation study}

\begin{table}[!t]
\small
\centering
\setlength{\tabcolsep}{2.0 pt}
\caption{A quantitative ablation study on the test set of C-GQA was conducted to verify the impact of different loss components on the performance of our model.}
\begin{tabular}{lcccccc}
\toprule
\multirow{2}{*}{\textbf{Loss term}}  & \multicolumn{6}{c}{C-GQA}          \\ 
\cmidrule(lr){2-7} 
& AUC &HM & Seen & Unseen & Attr. & Obj. \\ 
\midrule
$\mathcal{L}_{p}$  &3.90  &15.3  &33.4  &14.9  &17.8  &33.8    \\ 
$\mathcal{L}_{p}+\mathcal{L}_{o}$ &4.29 &16.1 &33.8 &15.3 & 8.0 &\textbf{43.9} \\ 
$\mathcal{L}_{p} + \mathcal{L}_{a}$ &4.16 &15.8 &33.4 &15.6 &\textbf{32.3}  &22.7   \\ 
$\mathcal{L}_{p}+\mathcal{L}_{a}+\mathcal{L}_{o}$ &4.60 &17.0 &33.3 &16.9  &15.5  &35.4   \\ 
$\mathcal{L}_{p} + \mathcal{L}_{a} + \mathcal{L}_{o} + \mathcal{L}_{cp} $ &\underline{4.68} &\underline{17.3} &\textbf{34.2} &\underline{17.0}  &15.9 &35.2  \\ 
$\mathcal{L}_{p}+ \mathcal{L}_{a} + \mathcal{L}_{o} + \mathcal{L}_{cp} + \mathcal{L}_{adv}$ &\textbf{4.89}  &\textbf{17.8}  &\underline{34.1}  &\textbf{17.8}   &\underline{22.4}   &\underline{35.7}   \\ 
\bottomrule
\end{tabular}
\label{tab:abalation}
\end{table}

As mentioned in Sec. \ref{sec:approach}, our PBadv method utilizes the technique of ensemble learning~\cite{2020_ensemble}, treating the final prediction score as a voting process that includes scores from both the conventional compositional branch \ding{195} and the disentangling branch \ding{192}, \ding{193}, and \ding{194} (shown in Fig. \ref{fig:main-arch}). Based on this, we proposed an improved adversarial training method tailored for the CZSL task and a data augmentation method based on object similarity. To verify the impact of each component on the final model classification performance, we conducted ablation studies on each branch and its corresponding loss components on the test set of C-GQA.

For clarity, we present all the ablation results in Table \ref{tab:abalation}. As each branch and module are progressively added, the classification model demonstrates a general upward trend in critical performance metrics, specifically the best AUC and HM. In other words, the overall performance of the model steadily improves with the inclusion of each component. The first row of Table \ref{tab:abalation} represents a model that uses only the compositional prediction branch, which reflects an earlier representative strategy~\cite{2021_cge}. As we add $\mathcal{L}_o$ and $\mathcal{L}_a$ separately, we observe an obvious improvement in the model's performance, with a noticeable increase in the best recognition accuracy of objects and attributes. This phenomenon can be interpreted from the perspective of the voting strategy in ensemble learning, the addition of parallel prediction branches increases the number of voters, and reduces the risk of over-fitting to some extent, thereby enhancing overall performance (\ie, AUC and HM).

It is worth noting that the target of the CZSL is to predict the compositional label of the given sample, rather than individual object or state. In other words, the prediction bias towards objects or states may not increase the classification accuracy for compositions. As shown by the results in the first and fourth rows of Table \ref{tab:abalation}, despite the overall performance of attributes and objects being not much different, the overall performance of the latter is significantly better than the former, with a performance improvement of 0.7\% in AUC and 1.7\% in HM. Differently, our model is that PBadv performs adversarial training not only on attributes and objects in the visual domain but also on the composed compositional features. Consequently, as shown in the last two rows of the table, adding the adversarial module improves both the overall performance of the model and the classification accuracy for attributes and objects. In summary, leveraging the technique of ensemble learning, each prediction branch contributes to the final performance of the model.

\begin{figure*}[!t]
\small
\centering
  \includegraphics[width=0.8\linewidth]{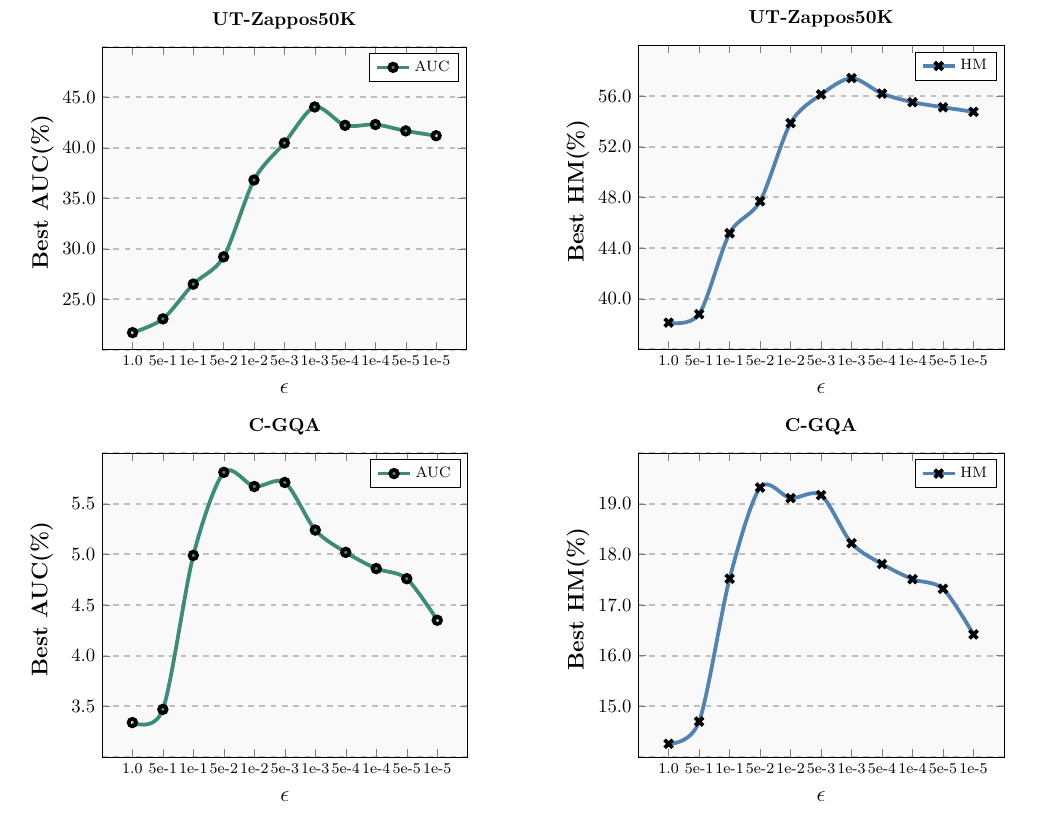}
    \caption
    {The influence of the temperature coefficient on UT-Zappos50K and C-GQA about the best AUC and HM. }
    \label{fig:hyper-parameter}
\end{figure*}

\subsection{Hyper-Parameter Analysis}
In our method, we utilize Cosine Similarity to measure the consistency between visual and textual features in the embedding space (as illustrated in Eqs. (\ref{equ:equ4}), (\ref{equ:equ5}), and (\ref{equ:equ6})). It is worth noting that temperature coefficient $\tau$ significantly influences the final performance of the model. To this end, we conduct a hyper-parameter analysis experiment to evaluate the performance of the model with various temperature coefficients on the validation sets of UT-Zappos50K and C-GQA. Based on empirical knowledge, we test a range of commonly used temperature coefficients, \{1.0,5e-1,1e-1,5e-2,1e-2,5e-3,1e-3,5e-4,1e-4,5e-5,1e-5\}, and from high to low, while maintaining the consistency of other model parts. As shown in Fig. \ref{fig:hyper-parameter}, we report the two representative metrics: the best AUC and HM.

From Fig. \ref{fig:hyper-parameter}, we observe that there exists an optimal temperature coefficient range for both the UT-Zappos50K and C-GQA datasets. For instance, when the temperature coefficient is within the range of [5e-2,1e-3], the model's performance on the UT-Zappos50K dataset improves significantly, achieving its peak on the validation set. Conversely, the model's evaluation metrics reach their maximum when the temperature coefficient lies between [1e-1,1e-2] on C-GQA. Based on it, we empirically selected a specific temperature coefficient for each dataset according to the model's performance on the validation set.

\subsection{Qualitative Analysis for Retrieval Experiments}
To intuitively demonstrate the actual performance of our model, we have drawn inspiration from previous outstanding works~\cite{2021_cge,2023learning_attention_disentangler} and conducted image-to-text, text-to-image, and primitive concept retrieval experiments on three widely used datasets.
\begin{figure*}[!th]
\small
\centering
  \includegraphics[width=1.0\linewidth]{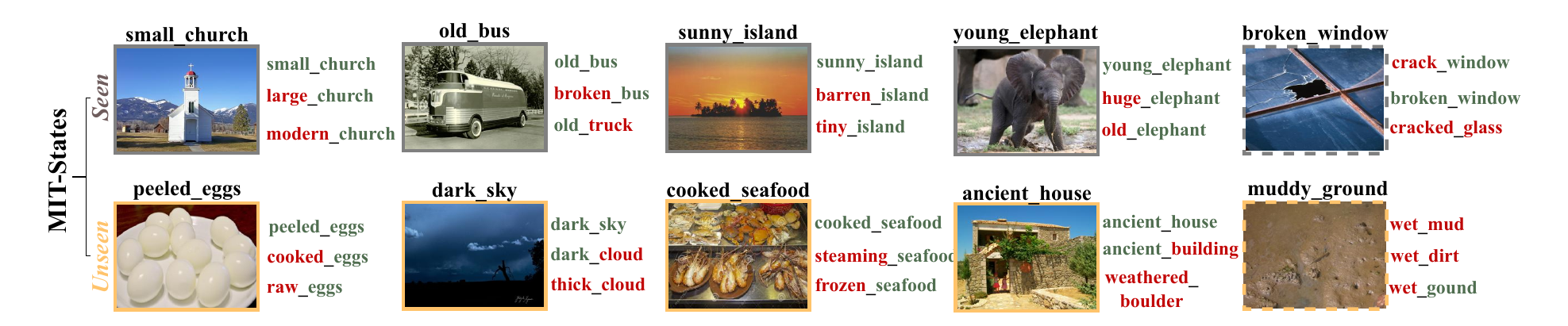}
    \caption
    {The qualitative results of image-to-text retrieval experiment on the test set of MIT-States.}
    \label{fig:retrieval_img2text}
\end{figure*}

\noindent\textbf{Image-to-Text Retrieval.}
As shown in Fig. \ref{fig:retrieval_img2text}, we first conducted image-to-text retrieval experiments on the test set of MIT-States. Specifically, we randomly selected several images from the seen compositions (top row) and unseen compositions (bottom row). The model outputs the top-3 score results most semantically similar to the pictures. For clarity, we placed the ground-truth labels in bold above each image. The top-3 predicted compositions are displayed to the right of the pictures, with correct predictions highlighted in green and incorrect ones in red.  Additionally, the images with incorrect predictions are outlined with dashed lines.

Thanks to our ensemble learning approach and the PBadv method, the model performs well in predicting compositions on the test set of MIT-States. Considering the variable and abstract nature of attributes in MIT-States, the accuracy in state recognition, even within seen compositions, is noticeably better than in object recognition. This can be attributed to two main reasons: Firstly, the training dataset for the model is relatively small in scale and has a significant attribute imbalance problem. Secondly, the current task setting treats the ground-truth label of the image as a single attribute-object pair, which does not align with real-life scenarios. In fact, objects tend to have multiple attributes, meaning a single object can be identified as various attribute-object pairs. For example, the model identifying $broken\_window$ as $cracked\_window$ is entirely reasonable, however, the dataset only annotates one compositional label. In summary, our image-to-text retrieval experiment demonstrates our method's ability to understand compositional semantics effectively.

\begin{figure*}[!t]
\small
\centering
  \includegraphics[width=1.0\linewidth]{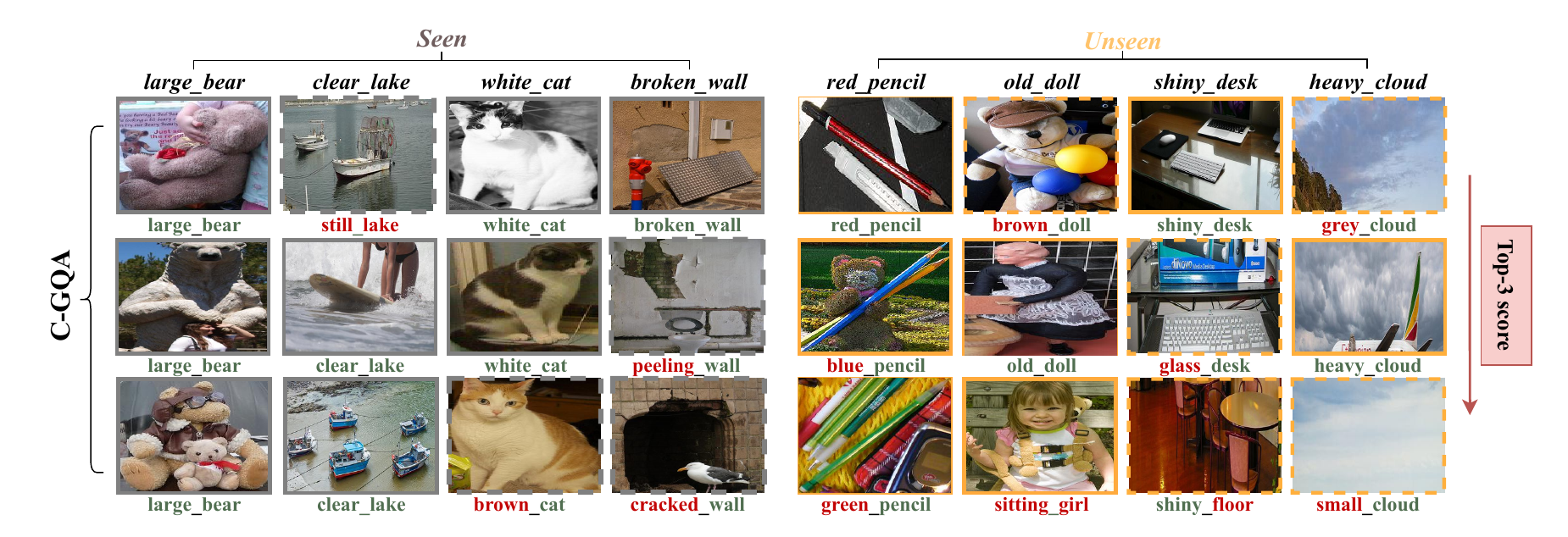}
    \caption
    {The qualitative results of text-to-image retrieval experiment on the test set of C-GQA.}
    \label{fig:txt2img}
\end{figure*}

\noindent\textbf{Text-to-Image Retrieval.}
Similarly, we conducted text-to-image retrieval experiments and reported the top-3 score retrieval results on the test set of the more challenging dataset: C-GQA, performing retrievals for both seen and unseen classes (As shown in Fig. \ref{fig:txt2img}). From Fig. \ref{fig:txt2img}, we can observe a phenomenon that the model performs significantly better on seen classes than unseen ones. This is a natural problem of the prediction bias towards seen classed in ZSL-related tasks. However, our model still exhibits a robust performance toward the perception of objects in both seen and unseen classes. It is worth noting that although image-to-text retrievals are not always entirely accurate, most of the results are still reasonable. For instance, it might identify $clear\_lake$ as $still\_lake$ or $heavy\_cloud$ as $grey\_cloud$. The image-to-text and text-to-image retrieval experiments demonstrate the ability of our model to maintain the semantic consistency alignment in the embedding common space.

\begin{figure*}[!t]
\small
\centering
  \includegraphics[width=1.0\linewidth]{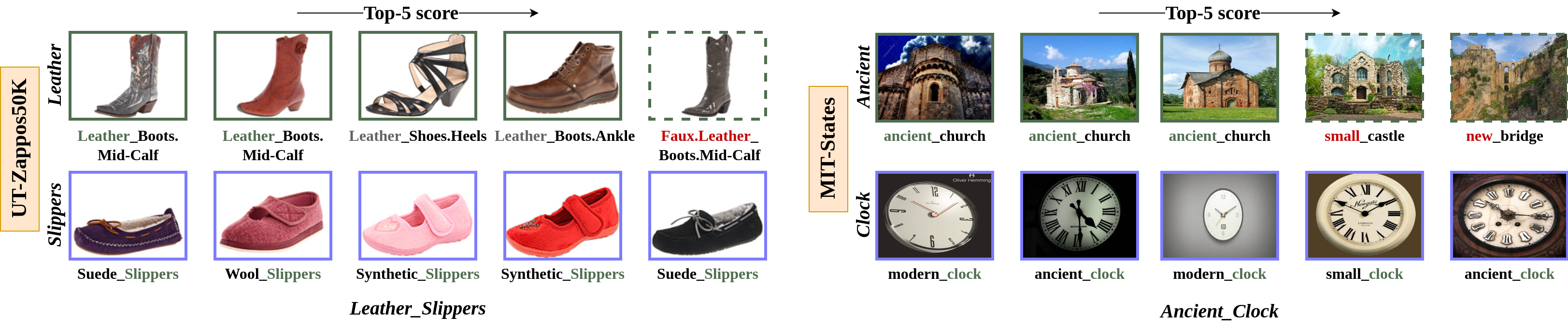}
    \caption
    {The qualitative results of primitive concept retrieval experiment on both test sets of UT-Zappos50K and MIT-States.}
    \label{fig:concept_exp}
\end{figure*}

\noindent\textbf{Visual Concept Retrieval.}
Considering that the primitives disentangling and semantic preservation are crucial in the decomposing and composing architecture, we also conducted retrieval experiments for attribute and object primitives and reported the top-5 retrieval results. Specifically, we randomly selected an unseen composition from the UT-Zappos50K and MIT-States datasets and performed retrieval for the primitive of attribute and object.

From Fig. \ref{fig:concept_exp}, we can observe that the model achieves highly accurate retrieval results for both attributes and objects on the UT-Zappos50K dataset. The only retrieval failure occurred when $Faux.Leather$ was retrieved as $Leather$. On the MIT-States dataset, the object retrieval performance is also outstanding, with only two wrong retrieval results in attribute retrieval. However, we find the two false retrieved images exhibit the features of $ancient$. We attribute this issue to the noisy annotations in the dataset rather than a fault of the model itself. These experiments demonstrate that our PBadv method enhances the model's ability to recognize attribute and object primitives.
\section{Conclusion}
We creatively proposed a method called Primitive-Based
Adversarial training (PBadv). By treating the contextual interactions in CZSL as adversarial learning, PBadv significantly improves the model's robustness to visual features with great discrepancy, achieving better overall performance. In addition, we designed a lightweight, plug-and-play data augmentation method (OS-OSP) to address the issues of the severe class imbalance and the under-fitting of hard compositions. Extensive quantitative and qualitative experiments illustrate the efficiency of our methods and achieve new state-of-the-art results on the three widely used datasets.

\begin{acks}
This work was supported in part by National Natural Science Foundation of China under the Grants 62371235 and 62072246, in part by Key Research and Development Plan of Jiangsu Province (Industry Foresight and Key Core Technology Project) under the Grant BE2023008-2.
\end{acks}

\bibliographystyle{ACM-Reference-Format}
\bibliography{sample-base}


\begin{thebibliography}{52}


\ifx \showCODEN    \undefined \def \showCODEN     #1{\unskip}     \fi
\ifx \showDOI      \undefined \def \showDOI       #1{#1}\fi
\ifx \showISBNx    \undefined \def \showISBNx     #1{\unskip}     \fi
\ifx \showISBNxiii \undefined \def \showISBNxiii  #1{\unskip}     \fi
\ifx \showISSN     \undefined \def \showISSN      #1{\unskip}     \fi
\ifx \showLCCN     \undefined \def \showLCCN      #1{\unskip}     \fi
\ifx \shownote     \undefined \def \shownote      #1{#1}          \fi
\ifx \showarticletitle \undefined \def \showarticletitle #1{#1}   \fi
\ifx \showURL      \undefined \def \showURL       {\relax}        \fi
\providecommand\bibfield[2]{#2}
\providecommand\bibinfo[2]{#2}
\providecommand\natexlab[1]{#1}
\providecommand\showeprint[2][]{arXiv:#2}

\bibitem[Atzmon et~al\mbox{.}(2020)]%
        {2020_mit_noise}
\bibfield{author}{\bibinfo{person}{Yuval Atzmon}, \bibinfo{person}{Felix Kreuk}, \bibinfo{person}{Uri Shalit}, {and} \bibinfo{person}{Gal Chechik}.} \bibinfo{year}{2020}\natexlab{}.
\newblock \showarticletitle{A causal view of compositional zero-shot recognition}. In \bibinfo{booktitle}{\emph{Advances in Neural Information Processing Systems}}, Vol.~\bibinfo{volume}{33}. \bibinfo{pages}{1462--1473}.
\newblock


\bibitem[Ba et~al\mbox{.}(2016)]%
        {2016_layernorm}
\bibfield{author}{\bibinfo{person}{Jimmy~Lei Ba}, \bibinfo{person}{Jamie~Ryan Kiros}, {and} \bibinfo{person}{Geoffrey~E Hinton}.} \bibinfo{year}{2016}\natexlab{}.
\newblock \showarticletitle{Layer normalization}.
\newblock \bibinfo{journal}{\emph{arXiv preprint arXiv:1607.06450}} (\bibinfo{year}{2016}).
\newblock


\bibitem[Caron et~al\mbox{.}(2021)]%
        {2021_dino}
\bibfield{author}{\bibinfo{person}{Mathilde Caron}, \bibinfo{person}{Hugo Touvron}, \bibinfo{person}{Ishan Misra}, \bibinfo{person}{Herv{\'e} J{\'e}gou}, \bibinfo{person}{Julien Mairal}, \bibinfo{person}{Piotr Bojanowski}, {and} \bibinfo{person}{Armand Joulin}.} \bibinfo{year}{2021}\natexlab{}.
\newblock \showarticletitle{Emerging properties in self-supervised vision transformers}. In \bibinfo{booktitle}{\emph{Proceedings of the IEEE/CVF International Conference on Computer Vision}}. \bibinfo{pages}{9650--9660}.
\newblock


\bibitem[Deng et~al\mbox{.}(2009)]%
        {2009_imagenet}
\bibfield{author}{\bibinfo{person}{Jia Deng}, \bibinfo{person}{Wei Dong}, \bibinfo{person}{Richard Socher}, \bibinfo{person}{Li-Jia Li}, \bibinfo{person}{Kai Li}, {and} \bibinfo{person}{Li Fei-Fei}.} \bibinfo{year}{2009}\natexlab{}.
\newblock \showarticletitle{Imagenet: A large-scale hierarchical image database}. In \bibinfo{booktitle}{\emph{Proceedings of the IEEE Conference on Computer Vision and Pattern Recognition}}. \bibinfo{pages}{248--255}.
\newblock


\bibitem[Dong et~al\mbox{.}(2020)]%
        {2020_ensemble}
\bibfield{author}{\bibinfo{person}{Xibin Dong}, \bibinfo{person}{Zhiwen Yu}, \bibinfo{person}{Wenming Cao}, \bibinfo{person}{Yifan Shi}, {and} \bibinfo{person}{Qianli Ma}.} \bibinfo{year}{2020}\natexlab{}.
\newblock \showarticletitle{A survey on ensemble learning}.
\newblock \bibinfo{journal}{\emph{Frontiers of Computer Science}}  \bibinfo{volume}{14} (\bibinfo{year}{2020}), \bibinfo{pages}{241--258}.
\newblock


\bibitem[Dong et~al\mbox{.}(2018)]%
        {2018boost_adv}
\bibfield{author}{\bibinfo{person}{Yinpeng Dong}, \bibinfo{person}{Fangzhou Liao}, \bibinfo{person}{Tianyu Pang}, \bibinfo{person}{Hang Su}, \bibinfo{person}{Jun Zhu}, \bibinfo{person}{Xiaolin Hu}, {and} \bibinfo{person}{Jianguo Li}.} \bibinfo{year}{2018}\natexlab{}.
\newblock \showarticletitle{Boosting adversarial attacks with momentum}. In \bibinfo{booktitle}{\emph{Proceedings of the IEEE Conference on Computer Vision and Pattern Recognition}}. \bibinfo{pages}{9185--9193}.
\newblock


\bibitem[Dosovitskiy et~al\mbox{.}(2021)]%
        {2021_vit}
\bibfield{author}{\bibinfo{person}{Alexey Dosovitskiy}, \bibinfo{person}{Lucas Beyer}, \bibinfo{person}{Alexander Kolesnikov}, \bibinfo{person}{Dirk Weissenborn}, \bibinfo{person}{Xiaohua Zhai}, \bibinfo{person}{Thomas Unterthiner}, \bibinfo{person}{Mostafa Dehghani}, \bibinfo{person}{Matthias Minderer}, \bibinfo{person}{Georg Heigold}, \bibinfo{person}{Sylvain Gelly}, \bibinfo{person}{Jakob Uszkoreit}, {and} \bibinfo{person}{Neil Houlsby}.} \bibinfo{year}{2021}\natexlab{}.
\newblock \bibinfo{title}{An Image is Worth 16x16 Words: Transformers for Image Recognition at Scale}.
\newblock
\newblock
\showeprint[arxiv]{2010.11929}~[cs.CV]


\bibitem[Duan et~al\mbox{.}(2022)]%
        {2022_adv-gen}
\bibfield{author}{\bibinfo{person}{Mingxing Duan}, \bibinfo{person}{Kenli Li}, \bibinfo{person}{Jiayan Deng}, \bibinfo{person}{Bin Xiao}, {and} \bibinfo{person}{Qi Tian}.} \bibinfo{year}{2022}\natexlab{}.
\newblock \showarticletitle{A novel multi-sample generation method for adversarial attacks}.
\newblock \bibinfo{journal}{\emph{ACM Transactions on Multimedia Computing, Communications, and Applications (TOMM)}} \bibinfo{volume}{18}, \bibinfo{number}{4} (\bibinfo{year}{2022}), \bibinfo{pages}{1--21}.
\newblock


\bibitem[Fu et~al\mbox{.}(2023)]%
        {2023_styleadv}
\bibfield{author}{\bibinfo{person}{Yuqian Fu}, \bibinfo{person}{Yu Xie}, \bibinfo{person}{Yanwei Fu}, {and} \bibinfo{person}{Yu-Gang Jiang}.} \bibinfo{year}{2023}\natexlab{}.
\newblock \showarticletitle{Styleadv: Meta style adversarial training for cross-domain few-shot learning}. In \bibinfo{booktitle}{\emph{Proceedings of the IEEE/CVF Conference on Computer Vision and Pattern Recognition}}. \bibinfo{pages}{24575--24584}.
\newblock


\bibitem[Goodfellow et~al\mbox{.}(2014)]%
        {2014_fgsm}
\bibfield{author}{\bibinfo{person}{Ian~J Goodfellow}, \bibinfo{person}{Jonathon Shlens}, {and} \bibinfo{person}{Christian Szegedy}.} \bibinfo{year}{2014}\natexlab{}.
\newblock \showarticletitle{Explaining and harnessing adversarial examples}.
\newblock \bibinfo{journal}{\emph{arXiv preprint arXiv:1412.6572}} (\bibinfo{year}{2014}).
\newblock


\bibitem[Hao et~al\mbox{.}(2023)]%
        {2023learning_attention_disentangler}
\bibfield{author}{\bibinfo{person}{Shaozhe Hao}, \bibinfo{person}{Kai Han}, {and} \bibinfo{person}{Kwan-Yee~K Wong}.} \bibinfo{year}{2023}\natexlab{}.
\newblock \showarticletitle{Learning Attention as Disentangler for Compositional Zero-shot Learning}. In \bibinfo{booktitle}{\emph{Proceedings of the IEEE/CVF Conference on Computer Vision and Pattern Recognition}}. \bibinfo{pages}{15315--15324}.
\newblock


\bibitem[He et~al\mbox{.}(2016)]%
        {2016_resnet}
\bibfield{author}{\bibinfo{person}{Kaiming He}, \bibinfo{person}{Xiangyu Zhang}, \bibinfo{person}{Shaoqing Ren}, {and} \bibinfo{person}{Jian Sun}.} \bibinfo{year}{2016}\natexlab{}.
\newblock \showarticletitle{Deep residual learning for image recognition}. In \bibinfo{booktitle}{\emph{Proceedings of the IEEE Conference on Computer Vision and Pattern Recognition}}. \bibinfo{pages}{770--778}.
\newblock


\bibitem[Hu and Ma(2022)]%
        {cm_adv_2}
\bibfield{author}{\bibinfo{person}{Yanxu Hu} {and} \bibinfo{person}{Andy~J Ma}.} \bibinfo{year}{2022}\natexlab{}.
\newblock \showarticletitle{Adversarial feature augmentation for cross-domain few-shot classification}. In \bibinfo{booktitle}{\emph{European Conference on Computer Vision}}. Springer, \bibinfo{pages}{20--37}.
\newblock


\bibitem[Isola et~al\mbox{.}(2015)]%
        {2015_mitstates}
\bibfield{author}{\bibinfo{person}{Phillip Isola}, \bibinfo{person}{Joseph~J Lim}, {and} \bibinfo{person}{Edward~H Adelson}.} \bibinfo{year}{2015}\natexlab{}.
\newblock \showarticletitle{Discovering states and transformations in image collections}. In \bibinfo{booktitle}{\emph{Proceedings of the IEEE Conference on Computer Vision and Pattern Recognition}}. \bibinfo{pages}{1383--1391}.
\newblock


\bibitem[Jiang and Zhang(2024)]%
        {2024_jiang}
\bibfield{author}{\bibinfo{person}{Chenyi Jiang} {and} \bibinfo{person}{Haofeng Zhang}.} \bibinfo{year}{2024}\natexlab{}.
\newblock \showarticletitle{Revealing the Proximate Long-Tail Distribution in Compositional Zero-Shot Learning}. In \bibinfo{booktitle}{\emph{Proceedings of the AAAI Conference on Artificial Intelligence}}, Vol.~\bibinfo{volume}{38}. \bibinfo{pages}{2498--2506}.
\newblock


\bibitem[Karthik et~al\mbox{.}(2021)]%
        {2021_revisiting_visual_product}
\bibfield{author}{\bibinfo{person}{Shyamgopal Karthik}, \bibinfo{person}{Massimiliano Mancini}, {and} \bibinfo{person}{Zeynep Akata}.} \bibinfo{year}{2021}\natexlab{}.
\newblock \showarticletitle{Revisiting visual product for compositional zero-shot learning}. In \bibinfo{booktitle}{\emph{NeurIPS 2021 Workshop on Distribution Shifts: Connecting Methods and Applications}}.
\newblock


\bibitem[Karthik et~al\mbox{.}(2022)]%
        {2022_kgsp}
\bibfield{author}{\bibinfo{person}{Shyamgopal Karthik}, \bibinfo{person}{Massimiliano Mancini}, {and} \bibinfo{person}{Zeynep Akata}.} \bibinfo{year}{2022}\natexlab{}.
\newblock \showarticletitle{KG-SP: Knowledge Guided Simple Primitives for Open World Compositional Zero-Shot Learning}. In \bibinfo{booktitle}{\emph{Proceedings of the IEEE/CVF Conference on Computer Vision and Pattern Recognition}}. \bibinfo{pages}{9336--9345}.
\newblock


\bibitem[Kim et~al\mbox{.}(2023)]%
        {2023_cot}
\bibfield{author}{\bibinfo{person}{Hanjae Kim}, \bibinfo{person}{Jiyoung Lee}, \bibinfo{person}{Seongheon Park}, {and} \bibinfo{person}{Kwanghoon Sohn}.} \bibinfo{year}{2023}\natexlab{}.
\newblock \showarticletitle{Hierarchical Visual Primitive Experts for Compositional Zero-Shot Learning}. In \bibinfo{booktitle}{\emph{Proceedings of the IEEE/CVF International Conference on Computer Vision}}. \bibinfo{pages}{5675--5685}.
\newblock


\bibitem[Kingma and Ba(2014)]%
        {2014_adam}
\bibfield{author}{\bibinfo{person}{Diederik~P Kingma} {and} \bibinfo{person}{Jimmy Ba}.} \bibinfo{year}{2014}\natexlab{}.
\newblock \showarticletitle{Adam: A method for stochastic optimization}.
\newblock \bibinfo{journal}{\emph{arXiv preprint arXiv:1412.6980}} (\bibinfo{year}{2014}).
\newblock


\bibitem[Kurakin et~al\mbox{.}(2016)]%
        {2016_adv_def}
\bibfield{author}{\bibinfo{person}{Alexey Kurakin}, \bibinfo{person}{Ian Goodfellow}, {and} \bibinfo{person}{Samy Bengio}.} \bibinfo{year}{2016}\natexlab{}.
\newblock \showarticletitle{Adversarial machine learning at scale}.
\newblock \bibinfo{journal}{\emph{arXiv preprint arXiv:1611.01236}} (\bibinfo{year}{2016}).
\newblock


\bibitem[Lake(2014)]%
        {lake2014towards}
\bibfield{author}{\bibinfo{person}{Brenden~M Lake}.} \bibinfo{year}{2014}\natexlab{}.
\newblock \emph{\bibinfo{title}{Towards more human-like concept learning in machines: Compositionality, causality, and learning-to-learn}}.
\newblock \bibinfo{thesistype}{Ph.\,D. Dissertation}. \bibinfo{school}{Massachusetts Institute of Technology}.
\newblock


\bibitem[Li et~al\mbox{.}(2020)]%
        {2020_symnet}
\bibfield{author}{\bibinfo{person}{Yong-Lu Li}, \bibinfo{person}{Yue Xu}, \bibinfo{person}{Xiaohan Mao}, {and} \bibinfo{person}{Cewu Lu}.} \bibinfo{year}{2020}\natexlab{}.
\newblock \showarticletitle{Symmetry and group in attribute-object compositions}. In \bibinfo{booktitle}{\emph{Proceedings of the IEEE/CVF Conference on Computer Vision and Pattern Recognition}}. \bibinfo{pages}{11316--11325}.
\newblock


\bibitem[Liu et~al\mbox{.}(2016)]%
        {2016delv_adv}
\bibfield{author}{\bibinfo{person}{Yanpei Liu}, \bibinfo{person}{Xinyun Chen}, \bibinfo{person}{Chang Liu}, {and} \bibinfo{person}{Dawn Song}.} \bibinfo{year}{2016}\natexlab{}.
\newblock \showarticletitle{Delving into transferable adversarial examples and black-box attacks}.
\newblock \bibinfo{journal}{\emph{arXiv preprint arXiv:1611.02770}} (\bibinfo{year}{2016}).
\newblock


\bibitem[Lu et~al\mbox{.}(2023)]%
        {2023decomposed_soft}
\bibfield{author}{\bibinfo{person}{Xiaocheng Lu}, \bibinfo{person}{Song Guo}, \bibinfo{person}{Ziming Liu}, {and} \bibinfo{person}{Jingcai Guo}.} \bibinfo{year}{2023}\natexlab{}.
\newblock \showarticletitle{Decomposed soft prompt guided fusion enhancing for compositional zero-shot learning}. In \bibinfo{booktitle}{\emph{Proceedings of the IEEE/CVF Conference on Computer Vision and Pattern Recognition}}. \bibinfo{pages}{23560--23569}.
\newblock


\bibitem[Madry et~al\mbox{.}(2017)]%
        {2017_pgd}
\bibfield{author}{\bibinfo{person}{Aleksander Madry}, \bibinfo{person}{Aleksandar Makelov}, \bibinfo{person}{Ludwig Schmidt}, \bibinfo{person}{Dimitris Tsipras}, {and} \bibinfo{person}{Adrian Vladu}.} \bibinfo{year}{2017}\natexlab{}.
\newblock \showarticletitle{Towards deep learning models resistant to adversarial attacks}.
\newblock \bibinfo{journal}{\emph{arXiv preprint arXiv:1706.06083}} (\bibinfo{year}{2017}).
\newblock


\bibitem[Mancini et~al\mbox{.}(2021)]%
        {2021_ow_czsl}
\bibfield{author}{\bibinfo{person}{Massimiliano Mancini}, \bibinfo{person}{Muhammad~Ferjad Naeem}, \bibinfo{person}{Yongqin Xian}, {and} \bibinfo{person}{Zeynep Akata}.} \bibinfo{year}{2021}\natexlab{}.
\newblock \showarticletitle{Open world compositional zero-shot learning}. In \bibinfo{booktitle}{\emph{Proceedings of the IEEE/CVF Conference on Computer Vision and Pattern Recognition}}. \bibinfo{pages}{5222--5230}.
\newblock


\bibitem[Misra et~al\mbox{.}(2017)]%
        {2017_red_wine}
\bibfield{author}{\bibinfo{person}{Ishan Misra}, \bibinfo{person}{Abhinav Gupta}, {and} \bibinfo{person}{Martial Hebert}.} \bibinfo{year}{2017}\natexlab{}.
\newblock \showarticletitle{From red wine to red tomato: Composition with context}. In \bibinfo{booktitle}{\emph{Proceedings of the IEEE Conference on Computer Vision and Pattern Recognition}}. \bibinfo{pages}{1792--1801}.
\newblock


\bibitem[Moosavi-Dezfooli et~al\mbox{.}(2017)]%
        {2017uni_adv}
\bibfield{author}{\bibinfo{person}{Seyed-Mohsen Moosavi-Dezfooli}, \bibinfo{person}{Alhussein Fawzi}, \bibinfo{person}{Omar Fawzi}, {and} \bibinfo{person}{Pascal Frossard}.} \bibinfo{year}{2017}\natexlab{}.
\newblock \showarticletitle{Universal adversarial perturbations}. In \bibinfo{booktitle}{\emph{Proceedings of the IEEE Conference on Computer Vision and Pattern Recognition}}. \bibinfo{pages}{1765--1773}.
\newblock


\bibitem[Naeem et~al\mbox{.}(2021)]%
        {2021_cge}
\bibfield{author}{\bibinfo{person}{Muhammad~Ferjad Naeem}, \bibinfo{person}{Yongqin Xian}, \bibinfo{person}{Federico Tombari}, {and} \bibinfo{person}{Zeynep Akata}.} \bibinfo{year}{2021}\natexlab{}.
\newblock \showarticletitle{Learning graph embeddings for compositional zero-shot learning}. In \bibinfo{booktitle}{\emph{Proceedings of the IEEE/CVF Conference on Computer Vision and Pattern Recognition}}. \bibinfo{pages}{953--962}.
\newblock


\bibitem[Nagarajan and Grauman(2018)]%
        {2018_attributes_operators}
\bibfield{author}{\bibinfo{person}{Tushar Nagarajan} {and} \bibinfo{person}{Kristen Grauman}.} \bibinfo{year}{2018}\natexlab{}.
\newblock \showarticletitle{Attributes as operators: factorizing unseen attribute-object compositions}. In \bibinfo{booktitle}{\emph{Proceedings of the European Conference on Computer Vision}}. \bibinfo{pages}{169--185}.
\newblock


\bibitem[Nair and Hinton(2010)]%
        {2010_relu}
\bibfield{author}{\bibinfo{person}{Vinod Nair} {and} \bibinfo{person}{Geoffrey~E Hinton}.} \bibinfo{year}{2010}\natexlab{}.
\newblock \showarticletitle{Rectified linear units improve restricted boltzmann machines}. In \bibinfo{booktitle}{\emph{Proceedings of the International Conference on Machine Learning}}. \bibinfo{pages}{807--814}.
\newblock


\bibitem[Nayak et~al\mbox{.}(2023)]%
        {2023learning_composed_soft}
\bibfield{author}{\bibinfo{person}{Nihal~V. Nayak}, \bibinfo{person}{Peilin Yu}, {and} \bibinfo{person}{Stephen~H. Bach}.} \bibinfo{year}{2023}\natexlab{}.
\newblock \bibinfo{title}{Learning to Compose Soft Prompts for Compositional Zero-Shot Learning}.
\newblock
\newblock
\showeprint[arxiv]{2204.03574}~[cs.LG]


\bibitem[Pennington et~al\mbox{.}(2014)]%
        {2014glove}
\bibfield{author}{\bibinfo{person}{Jeffrey Pennington}, \bibinfo{person}{Richard Socher}, {and} \bibinfo{person}{Christopher~D Manning}.} \bibinfo{year}{2014}\natexlab{}.
\newblock \showarticletitle{Glove: Global vectors for word representation}. In \bibinfo{booktitle}{\emph{Proceedings of the 2014 Conference on Empirical Methods in Natural Language Processing}}. \bibinfo{pages}{1532--1543}.
\newblock


\bibitem[Radford et~al\mbox{.}(2021)]%
        {clip}
\bibfield{author}{\bibinfo{person}{Alec Radford}, \bibinfo{person}{Jong~Wook Kim}, \bibinfo{person}{Chris Hallacy}, \bibinfo{person}{Aditya Ramesh}, \bibinfo{person}{Gabriel Goh}, \bibinfo{person}{Sandhini Agarwal}, \bibinfo{person}{Girish Sastry}, \bibinfo{person}{Amanda Askell}, \bibinfo{person}{Pamela Mishkin}, \bibinfo{person}{Jack Clark}, {et~al\mbox{.}}} \bibinfo{year}{2021}\natexlab{}.
\newblock \showarticletitle{Learning transferable visual models from natural language supervision}. In \bibinfo{booktitle}{\emph{International Conference on Machine Learning}}. PMLR, \bibinfo{pages}{8748--8763}.
\newblock


\bibitem[Sagi and Rokach(2018)]%
        {2018_ensemble}
\bibfield{author}{\bibinfo{person}{Omer Sagi} {and} \bibinfo{person}{Lior Rokach}.} \bibinfo{year}{2018}\natexlab{}.
\newblock \showarticletitle{Ensemble learning: A survey}.
\newblock \bibinfo{journal}{\emph{Wiley Interdisciplinary Reviews: Data Mining and Knowledge Discovery}} \bibinfo{volume}{8}, \bibinfo{number}{4} (\bibinfo{year}{2018}), \bibinfo{pages}{e1249}.
\newblock


\bibitem[Saini et~al\mbox{.}(2022)]%
        {2022OADis}
\bibfield{author}{\bibinfo{person}{Nirat Saini}, \bibinfo{person}{Khoi Pham}, {and} \bibinfo{person}{Abhinav Shrivastava}.} \bibinfo{year}{2022}\natexlab{}.
\newblock \showarticletitle{Disentangling Visual Embeddings for Attributes and Objects}. In \bibinfo{booktitle}{\emph{Proceedings of the IEEE/CVF Conference on Computer Vision and Pattern Recognition}}. \bibinfo{pages}{13658--13667}.
\newblock


\bibitem[Srivastava et~al\mbox{.}(2014)]%
        {2014_dropout}
\bibfield{author}{\bibinfo{person}{Nitish Srivastava}, \bibinfo{person}{Geoffrey Hinton}, \bibinfo{person}{Alex Krizhevsky}, \bibinfo{person}{Ilya Sutskever}, {and} \bibinfo{person}{Ruslan Salakhutdinov}.} \bibinfo{year}{2014}\natexlab{}.
\newblock \showarticletitle{Dropout: a simple way to prevent neural networks from overfitting}.
\newblock \bibinfo{journal}{\emph{Journal of Machine Learning Research}} \bibinfo{volume}{15}, \bibinfo{number}{1} (\bibinfo{year}{2014}), \bibinfo{pages}{1929--1958}.
\newblock


\bibitem[Szegedy et~al\mbox{.}(2013)]%
        {2013intriguing}
\bibfield{author}{\bibinfo{person}{Christian Szegedy}, \bibinfo{person}{Wojciech Zaremba}, \bibinfo{person}{Ilya Sutskever}, \bibinfo{person}{Joan Bruna}, \bibinfo{person}{Dumitru Erhan}, \bibinfo{person}{Ian Goodfellow}, {and} \bibinfo{person}{Rob Fergus}.} \bibinfo{year}{2013}\natexlab{}.
\newblock \showarticletitle{Intriguing properties of neural networks}.
\newblock \bibinfo{journal}{\emph{arXiv preprint arXiv:1312.6199}} (\bibinfo{year}{2013}).
\newblock


\bibitem[Tram{\`e}r et~al\mbox{.}(2017)]%
        {2017_ensemble_adv}
\bibfield{author}{\bibinfo{person}{Florian Tram{\`e}r}, \bibinfo{person}{Alexey Kurakin}, \bibinfo{person}{Nicolas Papernot}, \bibinfo{person}{Ian Goodfellow}, \bibinfo{person}{Dan Boneh}, {and} \bibinfo{person}{Patrick McDaniel}.} \bibinfo{year}{2017}\natexlab{}.
\newblock \showarticletitle{Ensemble adversarial training: Attacks and defenses}.
\newblock \bibinfo{journal}{\emph{arXiv preprint arXiv:1705.07204}} (\bibinfo{year}{2017}).
\newblock


\bibitem[Wang and Deng(2021)]%
        {cm_adv_1}
\bibfield{author}{\bibinfo{person}{Haoqing Wang} {and} \bibinfo{person}{Zhi-Hong Deng}.} \bibinfo{year}{2021}\natexlab{}.
\newblock \showarticletitle{Cross-domain few-shot classification via adversarial task augmentation}.
\newblock \bibinfo{journal}{\emph{arXiv preprint arXiv:2104.14385}} (\bibinfo{year}{2021}).
\newblock


\bibitem[Wang et~al\mbox{.}({[n.\,d.]})]%
        {2024_adv-ap1}
\bibfield{author}{\bibinfo{person}{Hanrui Wang}, \bibinfo{person}{Shuo Wang}, \bibinfo{person}{Cunjian Chen}, \bibinfo{person}{Massimo Tistarelli}, {and} \bibinfo{person}{Zhe Jin}.} \bibinfo{year}{[n.\,d.]}\natexlab{}.
\newblock \showarticletitle{A Multi-task Adversarial Attack Against Face Authentication}.
\newblock \bibinfo{journal}{\emph{ACM Transactions on Multimedia Computing, Communications and Applications}} (\bibinfo{year}{[n.\,d.]}).
\newblock


\bibitem[Wang et~al\mbox{.}(2023)]%
        {2023_learning_conditional}
\bibfield{author}{\bibinfo{person}{Qingsheng Wang}, \bibinfo{person}{Lingqiao Liu}, \bibinfo{person}{Chenchen Jing}, \bibinfo{person}{Hao Chen}, \bibinfo{person}{Guoqiang Liang}, \bibinfo{person}{Peng Wang}, {and} \bibinfo{person}{Chunhua Shen}.} \bibinfo{year}{2023}\natexlab{}.
\newblock \showarticletitle{Learning Conditional Attributes for Compositional Zero-Shot Learning}. In \bibinfo{booktitle}{\emph{Proceedings of the IEEE/CVF Conference on Computer Vision and Pattern Recognition}}. \bibinfo{pages}{11197--11206}.
\newblock


\bibitem[Wang et~al\mbox{.}(2021)]%
        {2021meta_adv}
\bibfield{author}{\bibinfo{person}{Ren Wang}, \bibinfo{person}{Kaidi Xu}, \bibinfo{person}{Sijia Liu}, \bibinfo{person}{Pin-Yu Chen}, \bibinfo{person}{Tsui-Wei Weng}, \bibinfo{person}{Chuang Gan}, {and} \bibinfo{person}{Meng Wang}.} \bibinfo{year}{2021}\natexlab{}.
\newblock \showarticletitle{On fast adversarial robustness adaptation in model-agnostic meta-learning}.
\newblock \bibinfo{journal}{\emph{arXiv preprint arXiv:2102.10454}} (\bibinfo{year}{2021}).
\newblock


\bibitem[Xu et~al\mbox{.}(2024)]%
        {2024_clip_gipcol}
\bibfield{author}{\bibinfo{person}{Guangyue Xu}, \bibinfo{person}{Joyce Chai}, {and} \bibinfo{person}{Parisa Kordjamshidi}.} \bibinfo{year}{2024}\natexlab{}.
\newblock \showarticletitle{GIPCOL: Graph-Injected Soft Prompting for Compositional Zero-Shot Learning}. In \bibinfo{booktitle}{\emph{Proceedings of the IEEE/CVF Winter Conference on Applications of Computer Vision}}. \bibinfo{pages}{5774--5783}.
\newblock


\bibitem[Xu et~al\mbox{.}(2023)]%
        {2023a2sc}
\bibfield{author}{\bibinfo{person}{Yikun Xu}, \bibinfo{person}{Xingxing Wei}, \bibinfo{person}{Pengwen Dai}, {and} \bibinfo{person}{Xiaochun Cao}.} \bibinfo{year}{2023}\natexlab{}.
\newblock \showarticletitle{A2SC: Adversarial Attacks on Subspace Clustering}.
\newblock \bibinfo{journal}{\emph{ACM Transactions on Multimedia Computing, Communications and Applications}} \bibinfo{volume}{19}, \bibinfo{number}{6} (\bibinfo{year}{2023}), \bibinfo{pages}{1--23}.
\newblock


\bibitem[Yang et~al\mbox{.}(2020)]%
        {2020_hierarchical_decompostition}
\bibfield{author}{\bibinfo{person}{Muli Yang}, \bibinfo{person}{Cheng Deng}, \bibinfo{person}{Junchi Yan}, \bibinfo{person}{Xianglong Liu}, {and} \bibinfo{person}{Dacheng Tao}.} \bibinfo{year}{2020}\natexlab{}.
\newblock \showarticletitle{Learning unseen concepts via hierarchical decomposition and composition}. In \bibinfo{booktitle}{\emph{Proceedings of the IEEE/CVF Conference on Computer Vision and Pattern Recognition}}. \bibinfo{pages}{10248--10256}.
\newblock


\bibitem[Yang et~al\mbox{.}(2023)]%
        {2022_decomposable_causal}
\bibfield{author}{\bibinfo{person}{Muli Yang}, \bibinfo{person}{Chenghao Xu}, \bibinfo{person}{Aming Wu}, {and} \bibinfo{person}{Cheng Deng}.} \bibinfo{year}{2023}\natexlab{}.
\newblock \showarticletitle{A Decomposable Causal View of Compositional Zero-Shot Learning}.
\newblock \bibinfo{journal}{\emph{IEEE Transactions on Multimedia}}  \bibinfo{volume}{25} (\bibinfo{year}{2023}), \bibinfo{pages}{5892--5902}.
\newblock


\bibitem[Yu and Grauman(2014)]%
        {2014_utzappos}
\bibfield{author}{\bibinfo{person}{Aron Yu} {and} \bibinfo{person}{Kristen Grauman}.} \bibinfo{year}{2014}\natexlab{}.
\newblock \showarticletitle{Fine-grained visual comparisons with local learning}. In \bibinfo{booktitle}{\emph{Proceedings of the IEEE Conference on Computer Vision and Pattern Recognition}}. \bibinfo{pages}{192--199}.
\newblock


\bibitem[Zeng et~al\mbox{.}(2022)]%
        {2022_adv_moment}
\bibfield{author}{\bibinfo{person}{Yawen Zeng}, \bibinfo{person}{Da Cao}, \bibinfo{person}{Shaofei Lu}, \bibinfo{person}{Hanling Zhang}, \bibinfo{person}{Jiao Xu}, {and} \bibinfo{person}{Zheng Qin}.} \bibinfo{year}{2022}\natexlab{}.
\newblock \showarticletitle{Moment is important: Language-based video moment retrieval via adversarial learning}.
\newblock \bibinfo{journal}{\emph{ACM Transactions on Multimedia Computing, Communications, and Applications (TOMM)}} \bibinfo{volume}{18}, \bibinfo{number}{2} (\bibinfo{year}{2022}), \bibinfo{pages}{1--21}.
\newblock


\bibitem[Zhang et~al\mbox{.}(2022)]%
        {2022_IVR}
\bibfield{author}{\bibinfo{person}{Tian Zhang}, \bibinfo{person}{Kongming Liang}, \bibinfo{person}{Ruoyi Du}, \bibinfo{person}{Xian Sun}, \bibinfo{person}{Zhanyu Ma}, {and} \bibinfo{person}{Jun Guo}.} \bibinfo{year}{2022}\natexlab{}.
\newblock \showarticletitle{Learning invariant visual representations for compositional zero-shot learning}. In \bibinfo{booktitle}{\emph{European Conference on Computer Vision}}. Springer, \bibinfo{pages}{339--355}.
\newblock


\bibitem[Zhang et~al\mbox{.}(2019)]%
        {2019_cos-tmp}
\bibfield{author}{\bibinfo{person}{Xiao Zhang}, \bibinfo{person}{Rui Zhao}, \bibinfo{person}{Yu Qiao}, \bibinfo{person}{Xiaogang Wang}, {and} \bibinfo{person}{Hongsheng Li}.} \bibinfo{year}{2019}\natexlab{}.
\newblock \showarticletitle{Adacos: Adaptively scaling cosine logits for effectively learning deep face representations}. In \bibinfo{booktitle}{\emph{Proceedings of the IEEE/CVF Conference on Computer Vision and Pattern Recognition}}. \bibinfo{pages}{10823--10832}.
\newblock


\bibitem[Zheng et~al\mbox{.}(2024)]%
        {2024caila}
\bibfield{author}{\bibinfo{person}{Zhaoheng Zheng}, \bibinfo{person}{Haidong Zhu}, {and} \bibinfo{person}{Ram Nevatia}.} \bibinfo{year}{2024}\natexlab{}.
\newblock \showarticletitle{CAILA: Concept-Aware Intra-Layer Adapters for Compositional Zero-Shot Learning}. In \bibinfo{booktitle}{\emph{Proceedings of the IEEE/CVF Winter Conference on Applications of Computer Vision}}. \bibinfo{pages}{1721--1731}.
\newblock


\end{thebibliography}

\end{document}